\documentclass[11pt]{article}

\usepackage[final]{acl}

\usepackage{times}
\usepackage{latexsym}

\usepackage[T1]{fontenc}
\usepackage[utf8]{inputenc}

\usepackage{microtype}

\usepackage{inconsolata}

\usepackage{graphicx}

\usepackage{enumitem}
\usepackage{amsmath}   % align, \text, \mid, \ldots, \Bigl/\Biggl 等
\usepackage{amssymb} 
\usepackage{multirow}
\usepackage{graphicx}
\usepackage{float}
\usepackage{booktabs}
\usepackage{tabularx}
\usepackage[most]{tcolorbox}
\usepackage{xcolor}
\usepackage{tablefootnote}
\usepackage{colortbl}
\usepackage{cleveref}
\usepackage{subcaption}
\title{PersonaDual: Balancing Personalization and Objectivity \\via Adaptive Reasoning}

\author{
Xiaoyou Liu$^{\diamondsuit}$\thanks{Equal contribution.}, Xinyi Mou$^{\diamondsuit}$\footnotemark[1], Shengbin Yue$^{\diamondsuit}$, Liang Wang$^{\diamondsuit}$, \\
\textbf{Yuqing Wang$^\clubsuit$, Qiexiang Wang$^\clubsuit$, Tianrui Qin$^\clubsuit$, Zhongyu Wei$^{\diamondsuit}$$^{\spadesuit}$} \\
$^{\diamondsuit}$Fudan University \quad $^{\spadesuit}$Shanghai Innovation Institute \quad $^\clubsuit$OPPO \\
\texttt{xiaoyouliu25@m.fudan.edu.cn} \quad \texttt{zywei@fudan.edu.cn} \\
}

\begin{document}
\maketitle

\begin{abstract}
As users increasingly expect LLMs to align with their preferences, personalized information becomes valuable. However, personalized information can be a double-edged sword: it can improve interaction but may compromise objectivity and factual correctness, especially when it is misaligned with the question. To alleviate this problem, we propose PersonaDual, a framework that supports both general-purpose objective reasoning and personalized reasoning in a single model, and adaptively switches modes based on context. PersonaDual is first trained with SFT to learn two reasoning patterns, and then further optimized via reinforcement learning with our proposed DualGRPO to improve mode selection. Experiments on objective and personalized benchmarks show that PersonaDual preserves the benefits of personalization while reducing interference, achieving near interference-free performance and better leveraging helpful personalized signals to improve objective problem-solving.
\end{abstract}

\section{Introduction}

Large Language Models (LLMs) have increasingly become integral to delivering interactive and personalized user services \citep{salemi2024lamp}. To support personalization, mainstream model families including GPT \citep{achiam2023gpt}, Gemini\citep{team2023gemini,comanici2025gemini}, Llama 3\citep{grattafiori2024llama}, and Claude \citep{anthropic2024claude3} have incorporated memory mechanisms to retain and reuse personalized information across sessions. While such information can significantly enhance user satisfaction \citep{tan2023user}, it may also reduce objectivity and induce factual errors or biases \citep{akpinar2025s,wei2023simple,gupta2023bias}, especially when the user's persona is irrelevant or unaligned with the query.

\begin{figure}[t]
    \centering
    \includegraphics[width=\columnwidth]{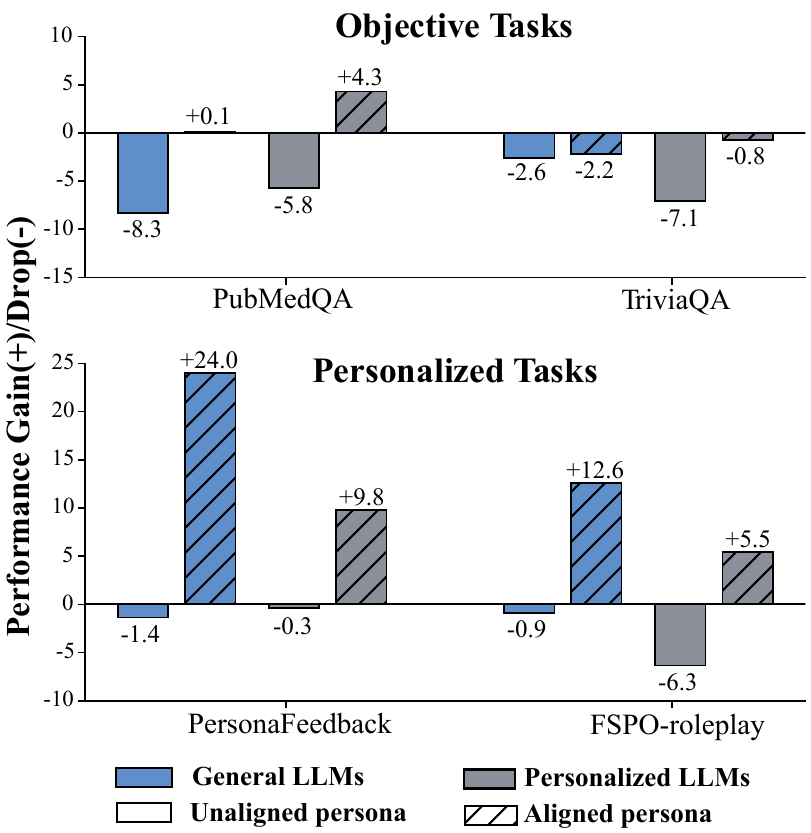}  % 图片路径
    % \vspace{-20pt}
    \caption{Effects of personalized information under different persona settings. We report the averaged results of DeepSeek-R1 \citep{guo2025deepseek} and Qwen3-30B-A3B-Thinking \citep{yang2025qwen3} for general models, and ALIGNXPERT-ICA \citep{li20251} and ALIGNXPERT-PBA \citep{li20251} for personalized models. Positive (negative) values indicate performance gains (drops) compared to the no-persona setting (indicated by the zero line).}
    \label{fig:introduction}
    % \vspace{-8pt} % 【关键】压缩Caption和正文之间的距离
\end{figure}

Prior studies on general-purpose LLMs show that personalized information can reduce factual accuracy \citep{akpinar2025s,wei2023simple}. However, these studies are largely descriptive and do not provide a systematic evaluation of personalization effects across different tasks. Meanwhile, existing personalization alignment methods, primarily realized through in-context learning \citep{wu2024understanding,salemi2025comparing,salemi2024optimization}, fine-tuning \citep{salemi2025comparing,finedemocratizing}, and reinforcement learning \citep{li20251}, have yielded measurable improvements in personalization performance. However their effects on logical reasoning and objective problem-solving remain underexplored.

To empirically investigate this duality, we evaluate both general-purpose and personalization-aligned models on objective factual and subjective personalized QA tasks under three settings: \textit{no persona}, \textit{unaligned persona}\footnote{For both objective and personalized tasks, unaligned personas are randomly sampled from PersonaHub \citep{ge2024scaling}}, and \textit{aligned persona}\footnote{For objective questions, aligned personas are generated by GPT-4o \citep{hurst2024gpt} to simulate a user whose background or interests are explicitly relevant to the question domain.}. Figure \ref{fig:introduction} summarizes the main results, with detailed analysis provided in Appendix \ref{app:introduction}). We have three findings. (1) Incorporating persona information generally reduces accuracy on objective factual QA, with performance drops of up to 8\% relative to the no-persona setting. (2) This degradation is most pronounced when personas are unaligned with the task, whereas aligned personas partially mitigate the effect and, in some cases, lead to modest accuracy gains of up to 4\%. (3) In contrast, for subjective personalization tasks, persona information substantially improves performance by facilitating preference inference, resulting in gains of approximately 10–20\%. Overall, these results suggest that the utility of personalized information is strongly task-dependent. While persona cues are beneficial for subjective personalization, they can interfere with objective reasoning when applied indiscriminately. This observation motivates a central research question: \textbf{How can LLMs adaptively determine when to incorporate personalized information, so as to balance personalization benefits with objective correctness?}

% ~\XY{I would rephrase it by: (1)~\textbf{Perturbation of Objective Reasoning}: Introducing any persona generally degrades factual QA accuracy by up to 8\%. (2)~\textbf{Modulation by Persona Alignment}: The performance drop is most severe with unaligned personas. However, aligned personas can mitigate or even reverse this trend, yielding accuracy gains of up to 4\% in some cases. (3)~\textbf{Essential Role in Subjective Tasks}: For subjective personalization tasks, persona cues are indispensable for inferring user preferences, typically leading to performance gains of 10–20\%.}

% Dual process theory \cite{kahneman2011thinking} is a well known psychological model suggesting that the human mind operates with two distinct thinking systems. Inspired by this theory, 

% ~\XY{@xiaoyou, add one more sentence introducing the mechanism of dual-process, e.g., they each handle different scenarios and can automatically switch based on xxx...}

To address this challenge, we draw inspiration from dual-process theory in psychology \citep{kahneman2011thinking}, which posits that human cognition relies on two distinct reasoning systems that are engaged depending on task demands. Motivated by this perspective, we propose \textbf{PersonaDual}, a framework that integrates general-purpose objective reasoning and personalized reasoning within a single model, and adaptively switches between these modes based on the input query and available personalized information.

In practice, we first construct a training dataset \textbf{PersonaDualData}, with a subset \textbf{PersonaDualData-SFT} containing reasoning trajectories in both objective and personalized modes. We further propose a dual-mode reasoning architecture and a two-stage training paradigm. In the first stage, supervised fine-tuning is used to train the model to produce outputs consistent with the two reasoning patterns. In the second stage, we introduce a reinforcement learning algorithm, \textbf{DualGRPO}, which optimizes adaptive mode selection conditioned on the user query and the available personalized information. We evaluate PersonaDual on both objective QA and subjective personalization benchmarks. Experimental results indicate that \textbf{PersonaDual substantially reduces the negative impact of unaligned personalized information on objective problem-solving}, achieving performance close to the no-persona setting. Furthermore, by adaptively switching between reasoning modes, \textbf{PersonaDual effectively leverages aligned personalized information, improving objective QA accuracy by nearly 3\%} on average compared to the no-persona baseline. %Our main contributions are as follows:

% ~\XY{DualGRPO was introduced without being defined. I would rephrase these sentences by: We further propose a dual-mode reasoning architecture and a two-stage training paradigm. First, supervised fine-tuning on the dataset teaches the model the two distinct reasoning patterns. Then, a new reinforcement learning algorithm DualGRPO is introduced to optimize the adaptive selection between modes conditioned on both user queries and available personalized information.}

%\begin{itemize}[label=\textbullet, leftmargin=*, itemsep=-2pt, topsep=4pt]
%\item
%    We systematically characterize the double-edged impact of personalized information, providing empirical evidence on how personas influence both personalization efficacy and objective accuracy, thus offering a valuable insights for future research.
%\item
%    We propose PersonaDual, a framework that balances subjective personalization and objective problem-solving under personalized information input. It is trained with a two-stage paradigm: supervised fine-tuning integrates general and personalized response modes, followed by reinforcement learning based on our proposed DualGRPO to enable adaptive mode switching.
%\item
%    Extensive experiments demonstrate the significant advantages of PersonaDual, achieving state-of-the-art performance on both objective and personalization tasks. For objective tasks, PersonaDual reduces interference from misaligned personas to within 0.6\% of an interference-free setting, and improves overall objective performance by 2.9\% under aligned personas.
%\end{itemize}

\section{Related Work}

\subsection{Impact of Personalization on Factuality}

As LLMs are increasingly used to serve diverse users and tasks, personalization in LLMs has become a major research focus. Existing personalization alignment methods typically adopt either context-based approaches \citep{packer2023memgpt,wu2024understanding,salemi2024optimization} or parameter-based alignment \citep{finedemocratizing,li20251}, substantially improving LLMs' ability to capture user-specific signals and produce personalized responses.

However, growing evidence suggests that personalized information can act as a strong prior in factual tasks and reduce reliability, shifting behavior from truth-seeking to agreement: preference-aligned models may exhibit sycophancy \citep{perez2023discovering}, override internal knowledge under misleading inputs and produce personalization-induced hallucinations \citep{wei2023simple}, or drift under incorrect user feedback \citep{chen2025self}. In addition, demographic attributes can trigger stereotypes and yield group-differentiated responses \citep{gallegos2024bias,sheng-etal-2021-societal}; pretraining biases can further amplify these risks in personalized settings \citep{liang2021towards}; and assigned personas can increase harmful or toxic outputs \citep{deshpande2023toxicity}. While recent work notes this "alignment tax" \citep{wang2025geometry} and attempts to mitigate it via architectural constraints, this approach largely relies  on static or manual triggers. Developing models that can intrinsically and adaptively decide when to leverage personalization remains an open question.

\begin{figure*}[t]
  \centering
  \includegraphics[width=\textwidth]{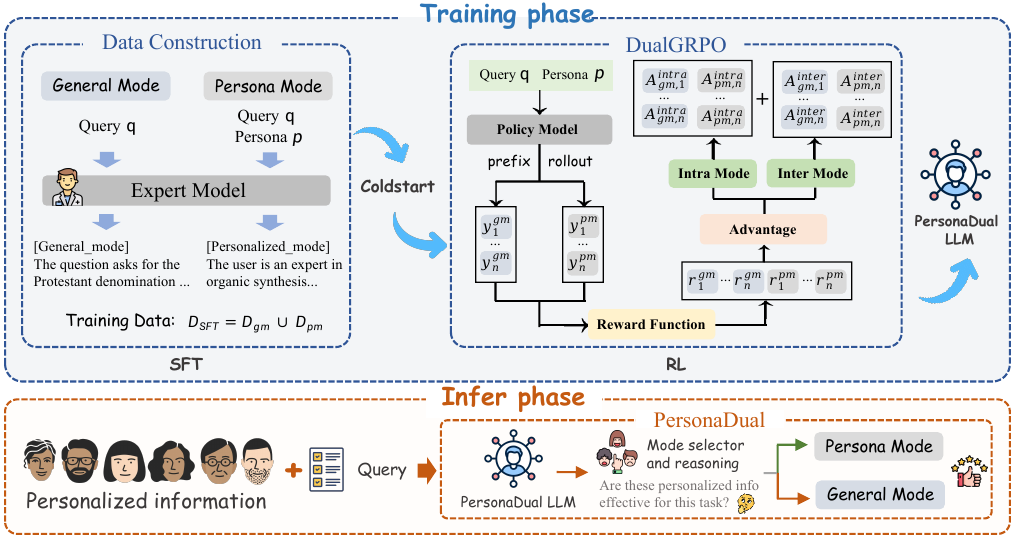}
  % \vspace{-24pt}
    \caption{{Overview of \textsc{PersonaDual} Framework.}
  \textsc{PersonaDual} equips a single LLM with two complementary response modes, objective and personalized reasoning, and learns to adaptively select between them based on the user query and available persona.
  Training follows a two-stage paradigm: supervised learning to disentangle the two reasoning behaviors, followed by \textsc{DualGRPO} to refine context-aware mode selection.
  This design captures the benefits of aligned persona cues while mitigating interference from irrelevant or unaligned personas, improving both objective correctness and personalization quality.}
  \label{fig:overview} 
  % \vspace{-8pt}
\end{figure*}

\subsection{Adaptive Inference and Dual-Process Architectures}

To address the diversity and complexity of real-world tasks, adaptive inference architectures have emerged that dynamically adjust computation paths or context strategies, ranging from routing-based mixtures \citep{fedus2022switch,jiang2024mixtral}, dynamic selection in model ensembles \citep{jiang2023llm}, to cascaded invocation via hierarchical reasoning \citep{chen2023frugalgpt}.

Recently, research has increasingly turned to cognitively inspired dual-process reasoning: drawing on the fast and slow thinking distinction, LLMs incorporate mechanisms to balance efficiency and quality, ranging from two-model setups for fast replies and deep planning \citep{tian2023duma} and fast/slow designs in human--AI collaboration \citep{xiao2025human,zhang2025leveraging} to multi-mode frameworks that learn adaptive selection to reduce chain-of-thought cost \citep{zhao2025frem,wang2025think,lou2025adacot}. While their work primarily targets efficiency, we instead aim to exploit the benefits of personalized information while mitigating its potential harms. Moreover, since most dual-mode models introduce modes via SFT and rely on RL for mode selection but often suffer from mode imbalance and slow selection updates \citep{fang2025thinkless,zhang2025adaptthink,gigerenzer2000adaptive,jiang2025think,li2025mixtureofvisualthoughtsexploringcontextadaptivereasoning}, we propose DualGRPO, which combines forced-prefix sampling and prefix-focused reinforcement to stabilize training and improve adaptive switching.

\section{Methods}

PersonaDual equips LLM with two response modes and allows it to autonomously switch its reasoning style according to the task context, thereby enabling adaptive personalized reasoning. The overall training procedure consists of two stages. We first integrate the two reasoning modes into the base model via SFT. Then we further propose DualGRPO, a reinforcement learning algorithm in the second training stage, enhancing the model’s ability to adaptively select response modes.

\subsection{Framework Overview}~\label{sec:method_overview}
% \XY{I think that elaborating on the inference stage here before Eq.2 (and not even specifying it as the inference stage) and then going back to the training stage would cause a break and confusion in understanding. I think a better way is to start from the task definition (infer stage) and then introduce the model design and training schema. I have rewritten these paragraphs. You can find your previous version in the comments. @xiaoyou, plz change a symbol for mode p since you have already used p for personal information, and check all the equations carefully.}
Our goal is to equip an LLM with the ability to adaptively decide when to utilize personalized information when answering a user query. Formally, let $q$ denote the query and $p$ the user’s persona. The model is designed to autonomously execute a two-step process during inference, as illustrated in the lower part of Figure \ref{fig:overview}: (1) mode selection: determine whether to rely on $p$ by choosing a reasoning mode $m$, with $m \in \mathcal{M} = \{gm, pm\}$, which we abstract as a mode-selection policy
\begin{equation}
\sigma(m \mid q,p), \quad m \in \{gm,pm\}.
\label{eq:policydefine}
\end{equation}
(2) conditional generation: produce the final response $y$ conditioned on the chosen mode. Concretely, the system samples a mode $m$, \mbox{$m \sim \sigma(\cdot \mid q,p)$}, prepends the corresponding prefix $\texttt{pfx}_m \in \{\texttt{pfx}_{gm}, \texttt{pfx}_{pm}\}$, and then generates the final response
\begin{equation}
y = y(q,p,m) \sim \pi_{\theta}(\cdot \mid q,p,\texttt{pfx}_m).
\label{eq:conditionresponsedefine}
\end{equation} 

To realize this, we propose PersonaDual, which integrates both reasoning modes within a single model. A dedicated learnable prefix token explicitly activates each mode: $\texttt{pfx}_{gm} = [\texttt{General\_mode}],\text{ } \texttt{pfx}_{pm} = [\texttt{Personalized\_mode}]$. In this way, the answer distribution is implemented as
\begin{equation}
\pi_{\theta}(y \mid q,p,m) := \pi_{\theta}(y \mid q,p,\texttt{pfx}_m),
\label{eq:distributiondefine}
\end{equation}
where $\theta$ denotes model parameters, and $\texttt{pfx}_m$ is prepended as a sequence prefix to explicitly inject the mode choice into the generation process.

Based on this, our goal is to jointly learn an optimal selector $\sigma^{\ast}$, and a compatible generator $\pi$ that maximizes expected response quality under the real-world data distribution $\mathcal{D}$. Let $R(q,p,m,y)$ denote the reward function for the generated answer. The overall objective can be written as:
\begin{equation}
\scalebox{0.77}{$
\max_{\pi}\; \mathbb{E}_{(q,p)\sim \mathcal{D}}
\Bigl[
\mathbb{E}_{m\sim \sigma(\cdot \mid q,p)}
\mathbb{E}_{y\sim \pi(\cdot \mid q,p,m)}
\bigl[
R(q,p,m,y)
\bigr]
\Bigr].
$}
\label{eq:finalobjective}
\end{equation}

To achieve this, we design a two-stage training scheme, illustrated in the upper part of Figure \ref{fig:overview}, which we detail as follows.

\subsection{Stage 1: Mixed Mode Integration}

% ===previous version===
% To equip the backbone with the ability to execute the two modes described above, we first perform a supervised warm-up via SFT, as shown in the upper part on the left side of Figure \ref{fig:overview}. We use DeepSeek-R1 \citep{guo2025deepseek} as an expert model to construct a high quality dataset that mixes reasoning trajectories from both modes.

% \XY{To equip the model with the ability to execute the two distinct reasoning modes defined in Sec~\ref{sec:method_overview}, we first perform a supervised fine-tuning (SFT) warm-up, as shown in the left part of the training pipeline in Figure \ref{fig:overview}. This stage aims to instill stable and high-quality reasoning patterns for each mode, preparing the model for subsequent adaptive selection. In general, we leverage a strong expert model, e.g., DeepSeek-R1 \citep{guo2025deepseek}, to construct a high-quality PersonaDual-SFT dataset, which contains parallel reasoning trajectories for both modes.}

To equip the model with the ability to execute the two distinct reasoning modes defined in Sec~\ref{sec:method_overview}, we first perform a supervised fine-tuning (SFT) warm-up, as shown in the left part of the training pipeline in Figure \ref{fig:overview}. This stage aims to instill stable and high-quality reasoning patterns for each mode, preparing the model for subsequent adaptive selection. In general, we leverage a strong expert model, e.g., DeepSeek-R1 \citep{guo2025deepseek}, to construct a high-quality PersonaDualData-SFT dataset, which contains reasoning trajectories for both modes.

\paragraph{General Purpose Objective Reasoning Mode.}
% ===previous version===
% We construct prompts containing only the user query $q$, without any personalized information $p$, and ask the expert model to generate reasoning steps based on factual evidence and logical reasoning. These trajectories encourage the model to form a stable depersonalized reasoning paradigm under the prefix constraint $pfx_{gm}$, focusing on objectivity, verifiability, and logical consistency of the query itself.

% \XY{For this mode, we construct prompts containing only the user query $q$, deliberately omitting any personalized information $p$. The expert model is then instructed to generate reasoning steps grounded in factual evidence and logical deduction. These trajectories, conditioned on the prefix $pfx_{gm}$, establish a stable depersonalized reasoning paradigm that prioritizes objectivity, verifiability, and the intrinsic logical structure of the query.}

For this mode, we construct prompts containing only the user query $q$, where personalized information $p$ is deliberately omitted. The expert model is then instructed to generate reasoning steps grounded in factual evidence and logical deduction. These trajectories, conditioned on the prefix $\texttt{pfx}_{gm}$, establish a stable depersonalized reasoning paradigm that prioritizes objectivity, verifiability, and the intrinsic logical structure of the query.

\paragraph{Personalized Reasoning Mode.}
% ===previous version===
% We provide the expert model with $(q,p)$ and ask it to explicitly analyze attributes in the user persona, such as occupation, interests, and affective needs, and to incorporate them as effective signals in the reasoning and decision process. This construction ensures that, when triggered by the prefix $pfx_{pm}$, the model actively leverages $p$ to enhance the relevance and individual fit of the responses, rather than merely appending user information at a superficial level.

% \XY{For the personalized mode, we provide the expert model with the full context $(q,p)$ and ask it to explicitly analyze attributes within the user persona, such as occupation, interests, and affective needs, and to actively incorporate these attributes as contextual signals throughout its reasoning chain. These trajectories, activated by the prefix $pfx_{pm}$, enable the model learns to meaningfully leverage $p$ to enhance the relevance and individual fit of responses, moving beyond superficial mention of user information.}

For personalized mode, we provide the expert model with the full context $(q,p)$ and ask it to explicitly analyze attributes within the user persona, such as occupation, interests, and affective needs, and to actively incorporate these attributes as contextual signals throughout its reasoning chain. These trajectories, activated by the prefix $\texttt{pfx}_{pm}$, enable the model to meaningfully leverage $p$ to enhance the relevance and individual fit of responses, moving beyond superficial mention of user information.

% ~\XY{Through this mixed-mode SFT, the model learns to map each control prefix to a corresponding, specialized reasoning behavior, fulfilling the conditional generation definition in Eq.~\eqref{eq:distributiondefine}.}

Through this mixed-mode SFT, the model learns to map each control prefix to a corresponding, specialized reasoning behavior, fulfilling the conditional generation definition in Eq.~\eqref{eq:distributiondefine}.

\subsection{Stage 2: Adaptive Mode Selection}

After SFT equips the model with two reasoning modes, we employ reinforcement learning to learn adaptive mode switching. To handle diverse task scenarios and personalized information, we propose DualGRPO (as shown in the top-right of Figure \ref{fig:overview}), an extension of GRPO designed for mode selection, featuring forced prefix sampling and inter-mode and intra-mode advantage calculation.

\paragraph{Prefix Forced Sampling.}
% ===previous version===
% After the SFT cold start, the model often exhibits a preference in selecting mode prefixes. For the same input $(q,p)$, the mode probabilities can become imbalanced, typically $\pi_{\theta}(pfx_{gm} \mid q,p) \neq \pi_{\theta}(pfx_{pm} \mid q,p)$. This preference can lead to an imbalanced number of sampled trajectories across two modes during reinforcement learning, which prevents a fair comparison of answer quality differences between modes under the same task and the same personalization configuration.

% \XY{After the SFT warm-up, we observe the model often exhibits a preference for one mode prefix over the other for a given input $(q,p)$, leading to imbalanced generation probabilities: $\pi_{\theta}(\texttt{pfx}_g \mid q, p) \gg \pi_{\theta}(\texttt{pfx}_p \mid q, p)$. This bias would cause an uneven number of sampled trajectories per mode during RL, thereby preventing a fair, controlled comparison of answer quality between the two modes under identical conditions. Such a comparison is a prerequisite for learning an optimal selection policy. Thus, we introduce a prefix-forced sampling strategy.  For each input $(q,p)$ during RL rollouts, we sample a total of $2n$ trajectories, explicitly enforcing a balanced distribution:}
% ===previous version===
% To remove this bias, in the sampling stage of RL, for each $(q,p)$ we sample a total of $2n$ trajectories and enforce that half of them use $pfx_{gm}$ and the other half use $pfx_{pm}$,

After the SFT warm-up, we observe the model often exhibits a preference for one mode prefix over the other for a given input $(q,p)$, leading to imbalanced generation probabilities: $\pi_{\theta}(\texttt{pfx}_{gm} \mid q, p) \gg \pi_{\theta}(\texttt{pfx}_{pm} \mid q, p)$. This bias would cause an uneven number of sampled trajectories per mode during RL, thereby preventing a fair, controlled comparison of answer quality between the two modes under identical conditions. Such a comparison is a prerequisite for learning an optimal selection policy. Thus, we introduce a prefix-forced sampling strategy.  For each input $(q,p)$ during RL rollouts, we sample a total of $2n$ trajectories, explicitly enforcing a balanced distribution:
% \begin{equation}
% \scalebox{0.65}{$
% \{y_{gm}^{(i)}\}_{i=1}^{n} \sim \pi_{\theta}(\cdot \mid q,p,\texttt{pfx}_{gm}), 
% \\
% \{y_{pm}^{(i)}\}_{i=1}^{n} \sim \pi_{\theta}(\cdot \mid q,p,\texttt{pfx}_{pm}).
% $}
% \label{eq:prefixsample}
% \end{equation}
\begin{equation}
\begin{aligned}
\{y_{gm}^{(i)}\}_{i=1}^{n} &\sim \pi_{\theta}(\cdot \mid q,p,\texttt{pfx}_{gm}), \\
\{y_{pm}^{(i)}\}_{i=1}^{n} &\sim \pi_{\theta}(\cdot \mid q,p,\texttt{pfx}_{pm}).
\end{aligned}
\label{eq:prefixsample}
\end{equation}
This provides a fair and controlled reference for learning the mode-selection policy $\sigma_{\phi}(m \mid q,p)$ in subsequent optimization.

\paragraph{Dual-Mode Advantage Decomposition.}

Standard GRPO typically computes relative advantages only within a single group of sampled responses and cannot directly reflect quality differences between two reasoning modes. Thus, with the balanced samples, our DualGRPO  computes advantages for policy updates using two levels of comparison. Concretely, for the same input $(q,p)$, let the sampled rewards under the general mode be $\mathcal{R}_{gm} = \{r_{gm}^{(i)}\}_{i=1}^{n}$ and similarly for the personalized mode $\mathcal{R}_{pm} = \{r_{pm}^{(i)}\}_{i=1}^{n}$. Let their means be $\mu_m = \frac{1}{n}\sum_{i=1}^{n} r_m^{(i)}$, we have:
% Thus, we decompose the advantage into two components, intra mode advantage and inter mode advantage. 
% \XY{Thus, with the balanced samples, our DualGRPO  computes advantages for policy updates using two levels of comparison. Concretely, } 
\begin{itemize}[label=\textbullet, leftmargin=2pt, itemsep=-2pt, topsep=6pt]
    \item \textbf{Intra-mode advantage}. We compute within-mode relative advantages by centering each sample reward with the mode-specific mean:
    \begin{equation}
        A_{m,i}^{\mathrm{intra}} = r_m^{(i)} - \mu_m, \ m \in \{gm, pm\}.
        \label{eq:intraadv}
    \end{equation}

    \item \textbf{Inter-mode advantage}. To reflect which mode is more suitable for the current input, we compare the two mode means and assign an opposing (zero-sum) signal across modes:
    \begin{equation}
    \scalebox{0.88}{$
        A_{m,i}^{\mathrm{inter}} = \mu_m - \mu_{\bar m}, \ \bar m \neq m,\; m,\bar m \in \{gm, pm\}.
        \label{eq:interadv}
    $}
    \end{equation}
    Intuitively, if $\mu_m > \mu_{\bar m}$, samples from mode $m$ receive a positive shift, while samples from the other mode are penalized, encouraging adaptive mode selection.

    \item \textbf{Composed advantage}. The final advantage adds the within-mode and cross-mode components:
    \begin{equation}
        A_{m,i} = A_{m,i}^{\mathrm{intra}} + A_{m,i}^{\mathrm{inter}}, \ m \in \{gm, pm\}.
        \label{eq:finaladv}
    \end{equation}
\end{itemize}

% This design provides two learning capabilities. It improves response quality within each individual mode, and it enables the model to learn adaptive selection of the better mode across different $(q,p)$ contexts. 

The combined advantage signal allows the policy to be updated toward choosing the mode that yields higher expected reward, thereby optimizing the joint objective in Eq.~\eqref{eq:finalobjective}.

% \XY{The combined advantage signal allows the policy to be updated toward choosing the mode that yields higher expected reward, thereby optimizing the joint objective in Eq.~\eqref{eq:finalobjective}. }

% \XY{TBD: @xiaoyou, is it possible to place this paragraph together with prefix forced sampling, as both are about the relevant operations on the prefix.}

Furthermore, to facilitate mode switching, we apply prefix strengthening by amplifying the advantage of prefix tokens by a factor $\beta>1$, namely $\beta \cdot A_{m,i}^{\texttt{pfx}}$, which in turn scales their contributions to the policy gradient. This mitigates the vanishing influence of early tokens in long reasoning trajectories, increases the effective learning signal for $p_m$, and accelerates the acquisition of a context-dependent mode selection policy.

% Furthermore, to facilitate mode switching, we apply prefix strengthening by up-weighting the policy-gradient terms of prefix tokens with an amplification factor $\beta>1$, mitigating the vanishing influence of early tokens in long reasoning trajectories. This increases the effective learning signal for $p_m$ and accelerates the acquisition of a context-dependent mode selection policy.

\section{Experimental Setup}

\label{sec:experimentalsetup}

% \subsection{Experimental Settings}

% \XY{This section details the experimental setup for evaluating PersonaDual. We first describe the modeling training, then introduce baselines for comparison, and finally outline the evaluation datasets and metrics.}

This section details the experimental setup for evaluating PersonaDual. We first describe the model training, then introduce baselines for comparison, and finally outline the evaluation datasets and metrics.

\paragraph{Training Details.} 
We implement the proposed PersonaDual using Qwen3-8B-Instruct\footnote{the non-thinking variant of Qwen3-8B} \citep{yang2025qwen3}. Our two-stage training utilizes a custom dataset PersonaDualData constructed by sampling from the general-purpose objective datasets UltraMedical \citep{ultramedical} and FLAN \citep{wei2022finetuned} alongside the personalization dataset AlignX \citep{li20251}. PersonaDualMode comprises 8,000 examples for SFT and 9,998 examples for RL. For objective tasks, we create two persona conditions: (1) Unaligned personas are randomly sampled from PersonaHub \citep{ge2024scaling}; (2) Aligned personas are relevant persona descriptions generated by GPT-4o \citep{hurst2024gpt} based on the question content. More details of experiments are provided in Table \ref{tab:training_data} and Appendix~\ref{app:Detailed Implementations}.

\paragraph{Baselines.} 
We select three categories of models as baselines for comparison: (1) \textbf{General purpose models}. \textbf{Qwen3-8B-Instruct} and \textbf{Llama-3.1-8B-Instruct} \citep{dubey2024llama} serve as strong instruction-following baselines. \textbf{CoT} is obtained through a two-stage training procedure with chain-of-thought supervision. \textbf{G-SFT-RL} is trained to improve the backbone’s objective reasoning capability. (2) \textbf{Personalization oriented models}. \textbf{Personal-Prompt} performs personalization alignment through prompting. \textbf{P-SFT-RL} is trained to enhance the backbone’s personalization capability. \textbf{ALIGNXPERT-ICA} \citep{li20251} and \textbf{ALIGNXPERT-PBA} \citep{li20251} perform large-scale persona alignment through in-context alignment and preference-based alignment, respectively. (3) \textbf{Dual-mode models}. \textbf{PersonaDual-Prompt} implements PersonaDual through prompting-based mode control. \textbf{PersonaDual-Router} realizes PersonaDual by employing an external router to switch between response modes.

\begin{table*}[!ht]
\centering
\small
\renewcommand{\arraystretch}{1.2}
\resizebox{\textwidth}{!}{
\begin{tabular}{l|cccccc|ccc}
\toprule
\multirow{2}{*}{\textbf{Model}} &
\multicolumn{6}{c|}{\textbf{Objective Acc. (Unalign./Align.)}} &
\multicolumn{3}{c}{\textbf{Personalized Acc. (Align.)}} \\
\cmidrule(lr){2-7}\cmidrule(lr){8-10}
& \textbf{PubMedQA} & \textbf{TriviaQA} & \textbf{MMLU-Pro} & \textbf{SuperGPQA} & \textbf{MATH500} & \textbf{Avg.}
& \textbf{P.FB.\tablefootnote{short for PersonaFeedback}} & \textbf{F.RP.\tablefootnote{short for FSPO-roleplay}} & \textbf{Avg.} \\
% FSPO-roleplay - > FSPO
\midrule

\multicolumn{10}{c}{\cellcolor[RGB]{247, 238, 199}{\textit{General purpose Models}}} \\
Qwen3-8B-Instruct      & 0.376/0.438 & 0.317/0.432 & \textbf{0.634}/\textbf{0.687} & 0.309/0.323 & 0.664/0.774 & 0.460/0.531 & 0.733 & 0.685 & 0.709 \\
Llama3.1-8B-Instruct   & 0.400/0.492 & 0.400/0.462 & 0.409/0.460 & 0.187/0.194 & 0.443/0.458 & 0.368/0.413 & 0.586 & 0.613 & 0.600 \\
CoT$^{\ast}$           & 0.445/0.534 & 0.438/0.500 & 0.606/0.637 & 0.317/0.316 & 0.764/0.776 & 0.514/0.553 & 0.740 & 0.787 & 0.764 \\
G-SFT-RL$^{\ast}$                & 0.491/0.538 & 0.432/0.504 & 0.617/0.643 & 0.321/0.330 & 0.772/0.778 & 0.527/0.559 & 0.658 & 0.720 & 0.689 \\
\midrule

\multicolumn{10}{c}{\cellcolor[RGB]{240, 207, 207}{\textit{Personalization oriented Models}}} \\
Personal-Prompt$^{\ast}$        & 0.359/0.446 & 0.252/0.349 & 0.488/0.539 & 0.287/0.313 & 0.353/0.386 & 0.348/0.407 & 0.741 & 0.695 & 0.718 \\
P-SFT-RL$^{\ast}$               & 0.349/0.390 & 0.351/0.446 & 0.576/0.653 & 0.310/0.330 & 0.669/0.696 & 0.451/0.503 & 0.738 & 0.762 & 0.750 \\
ALIGNXPERT-ICA$^{\ast}$         & 0.380/0.439 & 0.319/0.431 & 0.596/0.655 & 0.311/0.325 & 0.645/0.770 & 0.450/0.524 & 0.733 & 0.686 & 0.710 \\
ALIGNXPERT-PBA$^{\ast}$         & 0.371/0.437 & 0.321/0.431 & 0.592/0.653 & 0.318/0.332 & 0.659/0.768 & 0.452/0.524 & 0.731 & 0.684 & 0.708 \\
\midrule

\multicolumn{10}{c}{\cellcolor[RGB]{198, 230, 204}{\textit{Dual-mode Models}}} \\
PersonaDual-Prompt$^{\ast}$     & 0.348/0.433 & 0.335/0.431 & 0.576/0.630 & 0.296/0.330 & 0.733/0.756 & 0.458/0.516 & 0.746 & 0.735 & 0.741 \\
PersonaDual-Router$^{\ast}$     & 0.473/0.419 & 0.438/0.504 & 0.600/0.644 & 0.296/0.330 & 0.775/0.786 & 0.516/0.537 & 0.735 & 0.767 & 0.751 \\
PersonaDual$^{\ast}$            & \textbf{0.508}/\textbf{0.549} & \textbf{0.449}/\textbf{0.515} & 0.622/0.657
                      & \textbf{0.331}/\textbf{0.347} & \textbf{0.790}/\textbf{0.808} & \textbf{0.540}/\textbf{0.575}
                      & \textbf{0.747} & \textbf{0.799} & \textbf{0.773} \\
\midrule

No-Persona Upperbound             & 0.526 & 0.449 & 0.633 & 0.338 & 0.791 & 0.547 & -- & -- & -- \\
\bottomrule
\end{tabular}}
% \vspace{-8pt}
\caption{Performance on objective and personalized benchmarks. For objective datasets, each entry reports accuracy under Unaligned/Aligned persona settings. Personalized benchmarks are evaluated under the Aligned setting. Reported numbers are averaged over three runs. Avg. denotes the mean accuracy across datasets within each block. \textbf{Bold} indicates the best performance on each benchmark (evaluated under the corresponding setting). Models marked with ${\ast}$ are trained (or constructed via prompting) based on Qwen3-8B-Instruct.}
\label{tab:mainresult}
\end{table*}

\paragraph{Evaluation.}

We construct a comprehensive evaluation suite covering both objective and subjective scenarios, as summarized in Table \ref{tab:eval_benchmarks}. \textbf{(1) For objective tasks}, we include: PubMedQA \citep{jin2019pubmedqa}, TriviaQA \citep{joshi2017triviaqa}, MMLU-Pro \citep{wang2024mmlu}, SuperGPQA \citep{du2025supergpqa} and MATH500 \citep{hendrycks2021measuring}, covering multiple-choice question answering, open-domain factual QA, and mathematical problem solving. \textbf{(2) For subjective personalization}, we have PersonaFeedback \citep{tao2025personafeedback} and FSPO-roleplay \citep{singh2025fspo}, where models must select the preferred response from a pair of candidates based on a given persona. Most tasks are framed as classification tasks with structured outputs, allowing accuracy to be computed as the evaluation metric. For TriviaQA, we adopt the open-domain setting without provided evidence and evaluate answer correctness via exact match. For PubMedQA and MATH500, we employ GPT-4o-mini \citep{achiam2023gpt} as an automated judge to first extract the concise answer and then verify its correctness.

\section{Experiment Results}

% \XY{This section presents a comprehensive evaluation of PersonaDual, structured around several research questions. We begin with overall performance comparisons (RQ1–RQ3), followed by detailed ablation and mechanism analyses (RQ4–RQ5) to validate the design and interpret the behavior of PersonaDual.}

This section presents a comprehensive evaluation of PersonaDual, structured around several research questions. We begin with overall performance comparisons (RQ1–RQ3), followed by detailed ablation and mechanism analyses (RQ4–RQ6) to validate the design and interpret the behavior of PersonaDual.
% \begin{itemize}[label=\textbullet, leftmargin=*, itemsep=0pt, topsep=0pt]
% \item \textbf{RQ1}: Can PersonaDual handle the dual‑edged effects of personalized information?(Sec~\ref{sec:main_res})
% \item \textbf{RQ2}: Can PersonaDual achieve balanced excellence in both general and personalized reasoning? (Sec~\ref{sec:main_res})
% \item \textbf{RQ3}: Does PersonaDual's success stem from learning to switch modes appropriately? (Sec~\ref{sec:main_res})
% \item \textbf{RQ4}: Why is DualGRPO necessary for PersonaDual? (Sec~\ref{sec:rq4})
% \item \textbf{RQ5}: What signals does PersonaDual use for mode selection? (Sec~\ref{sec:rq5})
% \item \textbf{RQ6}: How robust is PersonaDual in multi‑turn dialogues? (Sec~\ref{sec:rq6})
% \end{itemize}

\subsection{Main Results}\label{sec:main_res}
Table~\ref{tab:mainresult} summarizes the overall performance of PersonaDual and all baselines on the benchmarks. Our analysis centers on three key research questions:
% ~\XY{Our analysis centers on three key research questions:}
% Based on these results, we have several following observations regarding the effectiveness of PersonaDual from different perspectives.
% \paragraph{RQ1: Can PersonaDual mitigate the double edged effects of personalized information?} 

\paragraph{RQ1: Can PersonaDual handle the dual‑edged effects of personalized information?}
PersonaDual successfully balances the influence of personalized information. When persona cues are \textbf{misaligned}, it effectively filters out interfering signals, achieving an average objective accuracy of 54.0\%, which closely matches the no-personalization upper bound (54.7\%). This shows that the model learns to screen and disregard irrelevant personalization. Meanwhile, when persona is \textbf{aligned} with the question, PersonaDual exploits these beneficial cues to improve factual accuracy, outperforming the no‑personalization upper bound by 2.8\% overall. These results indicate that our approach not only resists harmful interference but also actively leverages useful personalized signals to enhance answer quality.

\begin{figure}[t]
    \centering
    \begin{subfigure}[t]{0.52\columnwidth}
        \centering
        \includegraphics[width=\linewidth]{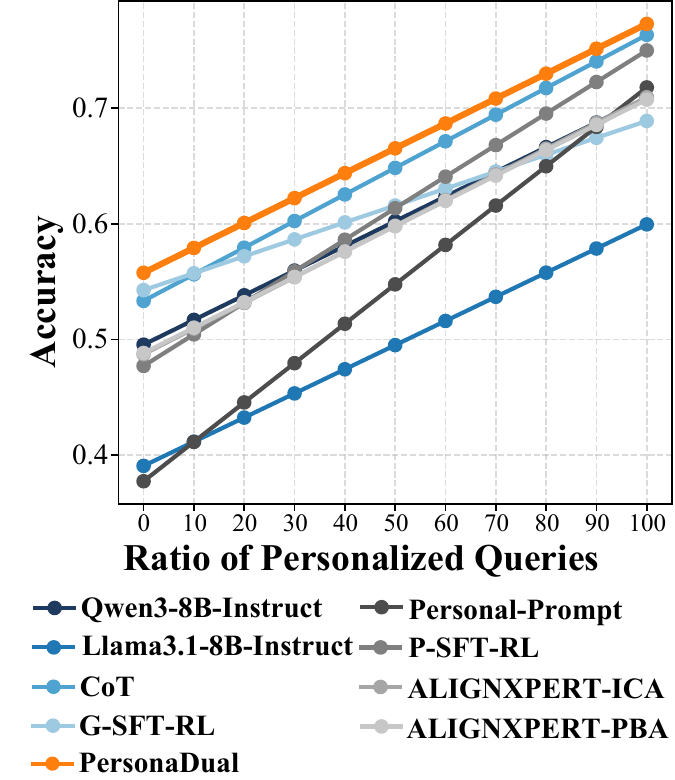}
        \caption{}
        \label{fig:combined:a}
    \end{subfigure}
    \hfill
    \begin{subfigure}[t]{0.43\columnwidth}
        \centering
        \includegraphics[width=\linewidth]{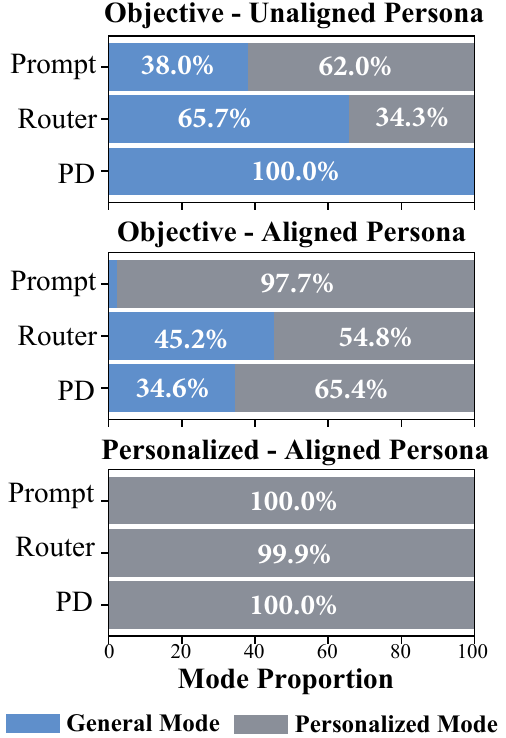}
        \caption{}
        \label{fig:combined:b}
    \end{subfigure}
    % \vspace{-6pt}
    \caption{(a) Performance comparison under mixed task settings with varying personalization ratios. (b) Comparison of mode proportions in dual-mode models. PD is short for PersonaDual.}
    \label{fig:combined}
\end{figure}

\paragraph{RQ2: Can PersonaDual achieve balanced excellence in both general and personalized reasoning?}
As shown in Table \ref{tab:mainresult}, general-purpose models, e.g., CoT and G-SFT-RL, achieve top rankings on objective tasks but perform poorly on personalized ones. Conversely, personalization‑oriented models exhibit the opposite trend, excelling in personalization at the cost of objective accuracy. \textbf{PersonaDual successfully bridges this gap: it matches the strong objective performance of general‑purpose models while simultaneously outperforming personalization‑focused models on personalized tasks,  thereby achieving the best overall balance}. To further assess robustness in realistic settings, we construct mixed test sets with varying proportions of personalized queries, as shown in Figure~\ref{fig:combined:a}. PersonaDual consistently outperforms both model families across all mixing ratios, validating its advantage in comprehensive scenarios.

\begin{table}[t]
\centering
\small
\renewcommand{\arraystretch}{1.15}
\resizebox{\columnwidth}{!}{
\begin{tabular}{m{1.2cm} m{2.6cm} c c c}
\toprule
\multirow{2}{*}{\textbf{Stage}}
& \multirow{2}{*}{\textbf{Model}}
& \multicolumn{2}{c}{\textbf{Objective}}
& \textbf{Personalized} \\
\cmidrule(lr){3-4} \cmidrule(lr){5-5}
&
& \textbf{Unalign.}
& \textbf{Align.}
& \textbf{Align.} \\
\midrule
\textbf{SFT}
& PersonaDual-SFT
& 52.4
& 55.3
& 74.4 \\
\midrule
\multirow{3}{*}{\textbf{RL}}
& PersonaDual
& \textbf{54.0}
& \textbf{57.5}
& \textbf{77.2} \\
& w/o DualAdv
& 53.3
& 55.9
& 75.5 \\
& \mbox{w/o DualAdv + PfxSmp}
& 53.6
& 56.1
& 76.9 \\
\bottomrule
\end{tabular}
}
% \vspace{-8pt}
\caption{Stage-wise comparison and RL ablations of PersonaDual.}
\label{tab:ablation}
\end{table}

\paragraph{RQ3: Does PersonaDual's success stem from learning to switch modes appropriately?}
The results in the Dual‑mode Models section of Table~\ref{tab:mainresult} show that PersonaDual achieves the best overall performance across all benchmarks. To understand this advantage, we analyze the models' mode‑selection behavior. As shown in Figure~\ref{fig:combined:b}, prompt‑based and router‑based baselines often activate the personalized mode inappropriately. For example, under misaligned persona settings, they still select personalized reasoning for 62.0\% and 34.3\% of objective instances, respectively. This indicates that hand‑designed prompts or heuristic rules fail to distinguish beneficial from harmful personalization signals. In contrast, \textbf{PersonaDual learns mode selection through training, resulting in more stable and reasonable switching across different contexts.}
\subsection{RQ4: Why is DualGRPO necessary for PersonaDual?}\label{sec:rq4}

% \paragraph{Compared to SFT} Overall, after RL training with DualGRPO, performance improves under both personalization configurations, including the aligned persona setting and the unaligned persona setting, as shown in the table \ref{tab:ablation}. In the aligned persona setting, our model’s average performance gain on objective test sets reaches 2.22\%, which is larger than the gain observed in the inconsistent persona setting, which is 1.6\%. This result suggests that DualGRPO in the RL stage primarily strengthens the model’s ability to identify beneficial personalized information and select the correct reasoning mode. This enables the model to better utilize helpful persona cues for objective questions, thereby improving objective task performance.
% \XY{\paragraph{Value over SFT} Table \ref{tab:ablation} shows that after RL training with DualGRPO, performance improves under both aligned and unaligned persona settings. The gain is more pronounced when personas are aligned (+2.2\% on objective tests) than when they are unaligned (+1.6\%). This indicates that DualGRPO specifically enhances the model’s ability to identify beneficial persona cues and select the correct reasoning mode, thereby leveraging helpful personalization to boost objective accuracy.}

\paragraph{Value over SFT} Table \ref{tab:ablation} shows that after RL training with DualGRPO, performance improves under both aligned and unaligned persona settings. The gain is more pronounced when personas are aligned (+2.2\% on objective tests) than when they are unaligned (+1.6\%). This indicates that DualGRPO specifically enhances the model’s ability to identify beneficial persona cues and select the correct reasoning mode, thereby leveraging helpful personalization to boost objective accuracy.

\paragraph{Ablation on DualGRPO Components} Table~\ref{tab:ablation} also shows that DualGRPO consistently outperforms ablated variants. Notably, we find that prefix‑forced sampling (PfxSmp) and dual‑mode advantage decomposition (DualAdv) are tightly coupled. Removing prefix constraints causes rollouts to collapse into a single mode, making mode‑aware credit assignment impossible. Conversely, without DualAdv, forced prefixes lack explicit credit guidance, preventing effective exploration. Thus, DualGRPO is necessary to enable flexible, context‑aware mode selection, which is critical when persona information varies.
\subsection{RQ5: What signals does PersonaDual use for mode selection?}\label{sec:rq5}
% \input{tables/personakeywords}

% \XY{\paragraph{Lexical Analysis via Predictive Tokens} To quantify how user information influences mode selection, we fit a logistic regression model on 2,500 randomly sampled instances, predicting the chosen mode from input tokens. By extracting the top‑50 tokens with the greatest predictive influence, as shown in Table \ref{tab:keywords}, we observe a clear pattern: occupational and professional terms, such as \textit{historian}, \textit{professor}, \textit{students}, and \textit{mathematician} appear most frequently. This indicates that descriptions of a user’s professional background serve as primary lexical signals for adaptive mode selection.}
\paragraph{Lexical Analysis via Predictive Tokens} To quantify how user information influences mode selection, we fit a logistic regression model on 2,500 randomly sampled instances, predicting the chosen mode from input tokens. By extracting the top‑30 tokens with the greatest predictive influence, as shown in Figure \ref{fig:keywords}, we observe a clear pattern: occupational and professional terms, such as \textit{historian}, \textit{professor}, \textit{students}, and \textit{mathematician} appear most frequently. This indicates that descriptions of a user’s professional background serve as primary lexical signals for adaptive mode selection.

\begin{figure}[!t]
    \centering
    \includegraphics[width=\columnwidth]{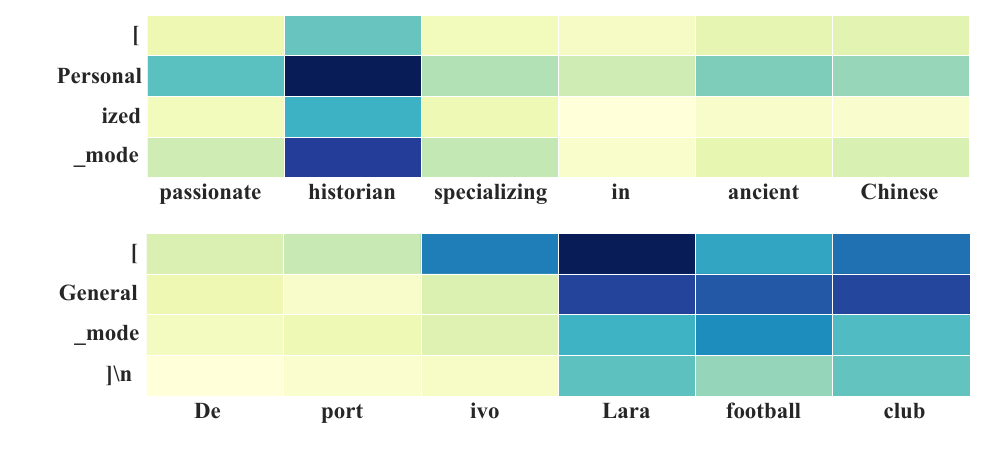}  % 图片路径
    % \vspace{-6pt} 
    \caption{Attention heatmap visualizing which persona keywords influence mode selection in PersonaDual. Darker colors indicate higher attention weights.}
    \label{fig:attention}
\end{figure}

\paragraph{Internal Validation through Attention Patterns} We further validate this finding by examining the model’s internal attention mechanisms. For the same query, we provide an aligned persona and an unaligned persona, then compare attention weights across persona tokens. As illustrated in Figure~\ref{fig:attention}, both the \texttt{[General\_mode]} and \texttt{[Personalized\_mode]} attend most strongly to tokens such as \textit{historian} and \textit{football} in the persona description. This consistency confirms that occupational and interest‑related tokens are central to the model’s reasoning process, directly linking these cues to the dual‑edged effect of personalization on objective and subjective tasks.

\begin{table}[t]
\centering
\small
\renewcommand{\arraystretch}{1.2}
\resizebox{\linewidth}{!}{
\begin{tabular}{llc} 
\toprule
\textbf{Task Type} & \textbf{Task Name} & \textbf{Deviation Ratio (\%)} \\
\midrule
\multirow{5}{*}{Objective}
& MATH500     & 60.53 \\
& MMLU-Pro    & 38.86 \\
& PubMedQA    & 68.63 \\
& SuperGPQA   & 30.73 \\
& TriviaQA    & 28.02 \\ 
\midrule
\multirow{2}{*}{Personalized}
& PersonaFeedback & 7.00 \\
& FSPO-roleplay   & 3.00 \\
\bottomrule
\end{tabular}}
% \vspace{-8pt}
\caption{Mode deviation ratio grouped by task type.}
\label{tab:mode_deviation}
\end{table}

\subsection{RQ6: How robust is PersonaDual in multi‑turn dialogues?}\label{sec:rq6}

% \XY{\paragraph{Analysis of Overall Deviation Pattern} To evaluate the robustness of PersonaDual in mixed question answering scenarios, we reconstruct the test set into two‑turn conversations. This simulates real‑world scenarios where users interleave different question types. Specifically, we construct two dialogue orders: (1) objective -> personalized and (2) personalized -> objective. We then define the mode‑change rate as whether the mode selected in the second turn matches the mode chosen for the same question in a single‑turn setting. The results reveal a clear asymmetry: when the first turn uses the personalized mode, PersonaDual maintains perfect consistency (0\% mode‑change rate). However, when starting with the objective mode, the mode‑change rate rises to 15.7\%. This indicates that switching from objective to personalized reasoning is a more challenging scenario for adaptive mode selection.}
\paragraph{Analysis of Overall Deviation Pattern} To evaluate the robustness of PersonaDual in mixed question answering scenarios, we reconstruct the test set into two‑turn conversations. This simulates real‑world scenarios where users interleave different question types. Specifically, we construct two dialogue orders: (1) general -> personalized and (2) personalized -> general. We then define the mode‑alignment rate as whether the mode selected in the second turn matches the mode chosen for the same question in a single‑turn setting. The results reveal a clear asymmetry: when the first turn uses the personalized mode, PersonaDual maintains perfect consistency (100\% mode‑alignment rate). However, when starting with the general mode, the mode‑alignment rate decreases to 84.3\%. This indicates that switching from objective to personalized reasoning is a more challenging scenario for adaptive mode selection.

\paragraph{Analysis of Deviation Source} To further understand the source of these deviations, we analyze the shifted cases by task type in Table \ref{tab:mode_deviation}. Mode shifts occur far more frequently on objective datasets than on personalization datasets. A plausible explanation is that, under the influence of a preceding objective turn, the model tends to rely more on superficial task‑type cues while paying less attention to the relevance between the personalized information and the current query. This heuristic can lead to a mode choice that differs from the single‑turn optimum. To address this, future work could incorporate multi‑turn training data, enabling more flexible mode switching and enhancing robustness in mixed dialogues.

\section{Conclusion}
To address the adverse impact of personalized information on the objectivity and factual accuracy of LLM responses, we propose PersonaDual, a framework that enables adaptive switching between general-purpose objective reasoning and personalized reasoning. PersonaDual unifies both reasoning modes within a single model via supervised fine-tuning, and further employs reinforcement learning with our proposed DualGRPO algorithm to enhance the model’s ability to select the appropriate reasoning mode conditioned on the input context. Extensive experiments demonstrate that PersonaDual consistently achieves strong performance across diverse personalization settings and task scenarios. It effectively mitigates interference caused by misaligned personalized information, achieving performance close to an interference-free setting, while simultaneously exploiting aligned personalized information to improve objective question answering by nearly 3\% over the no-personalization upper bound.

\section*{Limitations}

Our evaluation of personalization is limited by the availability of persona-related benchmarks, and currently relies on PersonaFeedback and FSPO-roleplay. Expanding to more diverse personalization benchmarks remains an important direction for future work. In addition, PersonaDual is mainly trained and evaluated in English, and extending the framework to multilingual settings is left for future exploration.

\section*{Ethics Statements}

We discuss two ethical considerations of our study. First, as described in Sec~\ref{sec:experimentalsetup} and Appendix~\ref{app:Detailed Implementations}, all personas and evaluation data are obtained from publicly available resources or are synthetically generated for research purposes. Hence, our experiments do not involve private user data and comply with standard privacy and security regulations. Second, our work studies LLM's performance under persona-conditioned inputs. A key risk is that persona cues may encode social identity signals, which can trigger stereotypes or yield group-differentiated responses. Such effects may amplify unfairness or produce overconfident yet incorrect outputs when personas are irrelevant or misleading. PersonaDual is explicitly designed to reduce this interference by adaptively selecting an objective reasoning mode when persona information is unhelpful.

% \section*{Acknowledgments}


\bibliography{custom}

\begin{thebibliography}{53}
\providecommand{\natexlab}[1]{#1}

\bibitem[{Achiam et~al.(2023)Achiam, Adler, Agarwal, Ahmad, Akkaya, Aleman, Almeida, Altenschmidt, Altman, Anadkat et~al.}]{achiam2023gpt}
Josh Achiam, Steven Adler, Sandhini Agarwal, Lama Ahmad, Ilge Akkaya, Florencia~Leoni Aleman, Diogo Almeida, Janko Altenschmidt, Sam Altman, Shyamal Anadkat, and 1 others. 2023.
\newblock Gpt-4 technical report.
\newblock \emph{arXiv preprint arXiv:2303.08774}.

\bibitem[{Akpinar et~al.(2025)Akpinar, Lee, Murdock, and Perona}]{akpinar2025s}
Nil-Jana Akpinar, Chia-Jung Lee, Vanessa Murdock, and Pietro Perona. 2025.
\newblock Who's asking? evaluating llm robustness to inquiry personas in factual question answering.
\newblock \emph{arXiv preprint arXiv:2510.12925}.

\bibitem[{Anthropic(2024)}]{anthropic2024claude3}
Anthropic. 2024.
\newblock \href {https://www-cdn.anthropic.com/de8ba9b01c9ab7cbabf5c33b80b7bbc618857627/Model_Card_Claude_3.pdf} {The claude 3 model family: Opus, sonnet, haiku}.
\newblock Technical report, Anthropic.

\bibitem[{Chen et~al.(2025)Chen, Huang, and Chen}]{chen2025self}
Chien~Hung Chen, Hen-Hsen Huang, and Hsin-Hsi Chen. 2025.
\newblock Self-augmented preference alignment for sycophancy reduction in llms.
\newblock In \emph{Proceedings of the 2025 Conference on Empirical Methods in Natural Language Processing}, pages 12390--12402.

\bibitem[{Chen et~al.(2023)Chen, Zaharia, and Zou}]{chen2023frugalgpt}
Lingjiao Chen, Matei Zaharia, and James Zou. 2023.
\newblock Frugalgpt: How to use large language models while reducing cost and improving performance.
\newblock \emph{arXiv preprint arXiv:2305.05176}.

\bibitem[{Comanici et~al.(2025)Comanici, Bieber, Schaekermann, Pasupat, Sachdeva, Dhillon, Blistein, Ram, Zhang, Rosen et~al.}]{comanici2025gemini}
Gheorghe Comanici, Eric Bieber, Mike Schaekermann, Ice Pasupat, Noveen Sachdeva, Inderjit Dhillon, Marcel Blistein, Ori Ram, Dan Zhang, Evan Rosen, and 1 others. 2025.
\newblock Gemini 2.5: Pushing the frontier with advanced reasoning, multimodality, long context, and next generation agentic capabilities.
\newblock \emph{arXiv preprint arXiv:2507.06261}.

\bibitem[{Deshpande et~al.(2023)Deshpande, Murahari, Rajpurohit, Kalyan, and Narasimhan}]{deshpande2023toxicity}
Ameet Deshpande, Vishvak Murahari, Tanmay Rajpurohit, Ashwin Kalyan, and Karthik Narasimhan. 2023.
\newblock Toxicity in chatgpt: Analyzing persona-assigned language models.
\newblock \emph{arXiv preprint arXiv:2304.05335}.

\bibitem[{Du et~al.(2025)Du, Yao, Ma, Wang, Zheng, Zhu, Liu, Liang, Jin, Wei et~al.}]{du2025supergpqa}
Xinrun Du, Yifan Yao, Kaijing Ma, Bingli Wang, Tianyu Zheng, King Zhu, Minghao Liu, Yiming Liang, Xiaolong Jin, Zhenlin Wei, and 1 others. 2025.
\newblock Supergpqa: Scaling llm evaluation across 285 graduate disciplines.
\newblock \emph{arXiv preprint arXiv:2502.14739}.

\bibitem[{Dubey et~al.(2024)Dubey, Jauhri, Pandey, Kadian, Al-Dahle, Letman, Mathur, Schelten, Yang, Fan et~al.}]{dubey2024llama}
Abhimanyu Dubey, Abhinav Jauhri, Abhinav Pandey, Abhishek Kadian, Ahmad Al-Dahle, Aiesha Letman, Akhil Mathur, Alan Schelten, Amy Yang, Angela Fan, and 1 others. 2024.
\newblock The llama 3 herd of models.
\newblock \emph{arXiv e-prints}, pages arXiv--2407.

\bibitem[{Fang et~al.(2025)Fang, Ma, and Wang}]{fang2025thinkless}
Gongfan Fang, Xinyin Ma, and Xinchao Wang. 2025.
\newblock Thinkless: Llm learns when to think.
\newblock \emph{arXiv preprint arXiv:2505.13379}.

\bibitem[{Fedus et~al.(2022)Fedus, Zoph, and Shazeer}]{fedus2022switch}
William Fedus, Barret Zoph, and Noam Shazeer. 2022.
\newblock Switch transformers: Scaling to trillion parameter models with simple and efficient sparsity.
\newblock \emph{Journal of Machine Learning Research}, 23(120):1--39.

\bibitem[{Gallegos et~al.(2024)Gallegos, Rossi, Barrow, Tanjim, Kim, Dernoncourt, Yu, Zhang, and Ahmed}]{gallegos2024bias}
Isabel~O Gallegos, Ryan~A Rossi, Joe Barrow, Md~Mehrab Tanjim, Sungchul Kim, Franck Dernoncourt, Tong Yu, Ruiyi Zhang, and Nesreen~K Ahmed. 2024.
\newblock Bias and fairness in large language models: A survey.
\newblock \emph{Computational Linguistics}, 50(3):1097--1179.

\bibitem[{Ge et~al.(2024)Ge, Chan, Wang, Yu, Mi, and Yu}]{ge2024scaling}
Tao Ge, Xin Chan, Xiaoyang Wang, Dian Yu, Haitao Mi, and Dong Yu. 2024.
\newblock Scaling synthetic data creation with 1,000,000,000 personas.
\newblock \emph{arXiv preprint arXiv:2406.20094}.

\bibitem[{Gigerenzer(2000)}]{gigerenzer2000adaptive}
Gerd Gigerenzer. 2000.
\newblock \emph{Adaptive thinking: Rationality in the real world}.
\newblock Oxford University Press, USA.

\bibitem[{Grattafiori et~al.(2024)Grattafiori, Dubey, Jauhri, Pandey, Kadian, Al-Dahle, Letman, Mathur, Schelten, Vaughan et~al.}]{grattafiori2024llama}
Aaron Grattafiori, Abhimanyu Dubey, Abhinav Jauhri, Abhinav Pandey, Abhishek Kadian, Ahmad Al-Dahle, Aiesha Letman, Akhil Mathur, Alan Schelten, Alex Vaughan, and 1 others. 2024.
\newblock The llama 3 herd of models.
\newblock \emph{arXiv preprint arXiv:2407.21783}.

\bibitem[{Guo et~al.(2025)Guo, Yang, Zhang, Song, Zhang, Xu, Zhu, Ma, Wang, Bi et~al.}]{guo2025deepseek}
Daya Guo, Dejian Yang, Haowei Zhang, Junxiao Song, Ruoyu Zhang, Runxin Xu, Qihao Zhu, Shirong Ma, Peiyi Wang, Xiao Bi, and 1 others. 2025.
\newblock Deepseek-r1: Incentivizing reasoning capability in llms via reinforcement learning.
\newblock \emph{arXiv preprint arXiv:2501.12948}.

\bibitem[{Gupta et~al.(2023)Gupta, Shrivastava, Deshpande, Kalyan, Clark, Sabharwal, and Khot}]{gupta2023bias}
Shashank Gupta, Vaishnavi Shrivastava, Ameet Deshpande, Ashwin Kalyan, Peter Clark, Ashish Sabharwal, and Tushar Khot. 2023.
\newblock Bias runs deep: Implicit reasoning biases in persona-assigned llms.
\newblock \emph{arXiv preprint arXiv:2311.04892}.

\bibitem[{Hendrycks et~al.(2021)Hendrycks, Burns, Kadavath, Arora, Basart, Tang, Song, and Steinhardt}]{hendrycks2021measuring}
Dan Hendrycks, Collin Burns, Saurav Kadavath, Akul Arora, Steven Basart, Eric Tang, Dawn Song, and Jacob Steinhardt. 2021.
\newblock Measuring mathematical problem solving with the math dataset.
\newblock \emph{arXiv preprint arXiv:2103.03874}.

\bibitem[{Hurst et~al.(2024)Hurst, Lerer, Goucher, Perelman, Ramesh, Clark, Ostrow, Welihinda, Hayes, Radford et~al.}]{hurst2024gpt}
Aaron Hurst, Adam Lerer, Adam~P Goucher, Adam Perelman, Aditya Ramesh, Aidan Clark, AJ~Ostrow, Akila Welihinda, Alan Hayes, Alec Radford, and 1 others. 2024.
\newblock Gpt-4o system card.
\newblock \emph{arXiv preprint arXiv:2410.21276}.

\bibitem[{Jiang et~al.(2024)Jiang, Sablayrolles, Roux, Mensch, Savary, Bamford, Chaplot, Casas, Hanna, Bressand et~al.}]{jiang2024mixtral}
Albert~Q Jiang, Alexandre Sablayrolles, Antoine Roux, Arthur Mensch, Blanche Savary, Chris Bamford, Devendra~Singh Chaplot, Diego de~las Casas, Emma~Bou Hanna, Florian Bressand, and 1 others. 2024.
\newblock Mixtral of experts.
\newblock \emph{arXiv preprint arXiv:2401.04088}.

\bibitem[{Jiang et~al.(2023)Jiang, Ren, and Lin}]{jiang2023llm}
Dongfu Jiang, Xiang Ren, and Bill~Yuchen Lin. 2023.
\newblock Llm-blender: Ensembling large language models with pairwise ranking and generative fusion.
\newblock \emph{arXiv preprint arXiv:2306.02561}.

\bibitem[{Jiang et~al.(2025)Jiang, Wu, Huang, Dong, Chi, Dong, Zhang, Lv, Cui, and Wei}]{jiang2025think}
Lingjie Jiang, Xun Wu, Shaohan Huang, Qingxiu Dong, Zewen Chi, Li~Dong, Xingxing Zhang, Tengchao Lv, Lei Cui, and Furu Wei. 2025.
\newblock Think only when you need with large hybrid-reasoning models.
\newblock \emph{arXiv preprint arXiv:2505.14631}.

\bibitem[{Jin et~al.(2019)Jin, Dhingra, Liu, Cohen, and Lu}]{jin2019pubmedqa}
Qiao Jin, Bhuwan Dhingra, Zhengping Liu, William Cohen, and Xinghua Lu. 2019.
\newblock Pubmedqa: A dataset for biomedical research question answering.
\newblock In \emph{Proceedings of the 2019 conference on empirical methods in natural language processing and the 9th international joint conference on natural language processing (EMNLP-IJCNLP)}, pages 2567--2577.

\bibitem[{Joshi et~al.(2017)Joshi, Choi, Weld, and Zettlemoyer}]{joshi2017triviaqa}
Mandar Joshi, Eunsol Choi, Daniel~S Weld, and Luke Zettlemoyer. 2017.
\newblock Triviaqa: A large scale distantly supervised challenge dataset for reading comprehension.
\newblock \emph{arXiv preprint arXiv:1705.03551}.

\bibitem[{Kahneman(2011)}]{kahneman2011thinking}
Daniel Kahneman. 2011.
\newblock \emph{Thinking, Fast and Slow}.
\newblock Farrar, Straus and Giroux, New York.

\bibitem[{Li et~al.(2025{\natexlab{a}})Li, Guan, Wu, Wu, and Yan}]{li20251}
Jia-Nan Li, Jian Guan, Songhao Wu, Wei Wu, and Rui Yan. 2025{\natexlab{a}}.
\newblock From 1,000,000 users to every user: Scaling up personalized preference for user-level alignment.
\newblock \emph{arXiv preprint arXiv:2503.15463}.

\bibitem[{Li et~al.(2025{\natexlab{b}})Li, Zhao, Zhang, Wang, Yao, Zhao, Song, Zheng, and Wei}]{li2025mixtureofvisualthoughtsexploringcontextadaptivereasoning}
Zejun Li, Yingxiu Zhao, Jiwen Zhang, Siyuan Wang, Yang Yao, Runzhou Zhao, Jun Song, Bo~Zheng, and Zhongyu Wei. 2025{\natexlab{b}}.
\newblock \href {https://arxiv.org/abs/2509.22746} {Mixture-of-visual-thoughts: Exploring context-adaptive reasoning mode selection for general visual reasoning}.
\newblock \emph{Preprint}, arXiv:2509.22746.

\bibitem[{Liang et~al.(2021)Liang, Wu, Morency, and Salakhutdinov}]{liang2021towards}
Paul~Pu Liang, Chiyu Wu, Louis-Philippe Morency, and Ruslan Salakhutdinov. 2021.
\newblock Towards understanding and mitigating social biases in language models.
\newblock In \emph{International conference on machine learning}, pages 6565--6576. PMLR.

\bibitem[{Lou et~al.(2025)Lou, Sun, Liang, Qu, Shen, Wang, Li, Yang, and Wu}]{lou2025adacot}
Chenwei Lou, Zewei Sun, Xinnian Liang, Meng Qu, Wei Shen, Wenqi Wang, Yuntao Li, Qingping Yang, and Shuangzhi Wu. 2025.
\newblock Adacot: Pareto-optimal adaptive chain-of-thought triggering via reinforcement learning.
\newblock \emph{arXiv preprint arXiv:2505.11896}.

\bibitem[{Packer et~al.(2023)Packer, Fang, Patil, Lin, Wooders, and Gonzalez}]{packer2023memgpt}
Charles Packer, Vivian Fang, Shishir~G Patil, Kevin Lin, Sarah Wooders, and Joseph~E Gonzalez. 2023.
\newblock Memgpt: Towards llms as operating systems.
\newblock \emph{CoRR}.

\bibitem[{Perez et~al.(2023)Perez, Ringer, Lukosiute, Nguyen, Chen, Heiner, Pettit, Olsson, Kundu, Kadavath et~al.}]{perez2023discovering}
Ethan Perez, Sam Ringer, Kamile Lukosiute, Karina Nguyen, Edwin Chen, Scott Heiner, Craig Pettit, Catherine Olsson, Sandipan Kundu, Saurav Kadavath, and 1 others. 2023.
\newblock Discovering language model behaviors with model-written evaluations.
\newblock In \emph{Findings of the association for computational linguistics: ACL 2023}, pages 13387--13434.

\bibitem[{Salemi et~al.(2024{\natexlab{a}})Salemi, Kallumadi, and Zamani}]{salemi2024optimization}
Alireza Salemi, Surya Kallumadi, and Hamed Zamani. 2024{\natexlab{a}}.
\newblock Optimization methods for personalizing large language models through retrieval augmentation.
\newblock In \emph{Proceedings of the 47th International ACM SIGIR Conference on Research and Development in Information Retrieval}, pages 752--762.

\bibitem[{Salemi et~al.(2024{\natexlab{b}})Salemi, Mysore, Bendersky, and Zamani}]{salemi2024lamp}
Alireza Salemi, Sheshera Mysore, Michael Bendersky, and Hamed Zamani. 2024{\natexlab{b}}.
\newblock Lamp: When large language models meet personalization.
\newblock In \emph{Proceedings of the 62nd Annual Meeting of the Association for Computational Linguistics (Volume 1: Long Papers)}, pages 7370--7392.

\bibitem[{Salemi and Zamani(2025)}]{salemi2025comparing}
Alireza Salemi and Hamed Zamani. 2025.
\newblock Comparing retrieval-augmentation and parameter-efficient fine-tuning for privacy-preserving personalization of large language models.
\newblock In \emph{Proceedings of the 2025 International ACM SIGIR Conference on Innovative Concepts and Theories in Information Retrieval (ICTIR)}, pages 286--296.

\bibitem[{Sheng et~al.(2021)Sheng, Chang, Natarajan, and Peng}]{sheng-etal-2021-societal}
Emily Sheng, Kai-Wei Chang, Prem Natarajan, and Nanyun Peng. 2021.
\newblock \href {https://doi.org/10.18653/v1/2021.acl-long.330} {Societal biases in language generation: Progress and challenges}.
\newblock In \emph{Proceedings of the 59th Annual Meeting of the Association for Computational Linguistics and the 11th International Joint Conference on Natural Language Processing (Volume 1: Long Papers)}, pages 4275--4293, Online. Association for Computational Linguistics.

\bibitem[{Singh et~al.(2025)Singh, Hsu, Hsu, Mitchell, Ermon, Hashimoto, Sharma, and Finn}]{singh2025fspo}
Anikait Singh, Sheryl Hsu, Kyle Hsu, Eric Mitchell, Stefano Ermon, Tatsunori Hashimoto, Archit Sharma, and Chelsea Finn. 2025.
\newblock Fspo: Few-shot preference optimization of synthetic preference data in llms elicits effective personalization to real users.
\newblock \emph{arXiv preprint arXiv:2502.19312}.

\bibitem[{Tan and Jiang(2023)}]{tan2023user}
Zhaoxuan Tan and Meng Jiang. 2023.
\newblock User modeling in the era of large language models: Current research and future directions.
\newblock \emph{arXiv preprint arXiv:2312.11518}.

\bibitem[{Tan et~al.(2024)Tan, Zeng, Tian, Liu, Yin, and Jiang}]{finedemocratizing}
Zhaoxuan Tan, Qingkai Zeng, Yijun Tian, Zheyuan Liu, Bing Yin, and Meng Jiang. 2024.
\newblock \href {https://doi.org/10.18653/v1/2024.emnlp-main.372} {Democratizing large language models via personalized parameter-efficient fine-tuning}.
\newblock In \emph{Proceedings of the 2024 Conference on Empirical Methods in Natural Language Processing}, pages 6476--6491, Miami, Florida, USA. Association for Computational Linguistics.

\bibitem[{Tao et~al.(2025)Tao, Zhu, Ding, Wang, Jiang, and Zhou}]{tao2025personafeedback}
Meiling Tao, Chenghao Zhu, Dongyi Ding, Tiannan Wang, Yuchen~Eleanor Jiang, and Wangchunshu Zhou. 2025.
\newblock Personafeedback: A large-scale human-annotated benchmark for personalization.
\newblock \emph{arXiv preprint arXiv:2506.12915}.

\bibitem[{Team et~al.(2023)Team, Anil, Borgeaud, Alayrac, Yu, Soricut, Schalkwyk, Dai, Hauth, Millican et~al.}]{team2023gemini}
Gemini Team, Rohan Anil, Sebastian Borgeaud, Jean-Baptiste Alayrac, Jiahui Yu, Radu Soricut, Johan Schalkwyk, Andrew~M Dai, Anja Hauth, Katie Millican, and 1 others. 2023.
\newblock Gemini: a family of highly capable multimodal models.
\newblock \emph{arXiv preprint arXiv:2312.11805}.

\bibitem[{Tian et~al.(2023)Tian, Chen, Liu, Liu, Zou, Chen, and Cui}]{tian2023duma}
Xiaoyu Tian, Liangyu Chen, Na~Liu, Yaxuan Liu, Wei Zou, Kaijiang Chen, and Ming Cui. 2023.
\newblock Duma: A dual-mind conversational agent with fast and slow thinking.
\newblock \emph{arXiv preprint arXiv:2310.18075}.

\bibitem[{Wang et~al.(2025)Wang, Li, Wang, Zhang, Xu, Wu, Huang, Yu, and Mao}]{wang2025think}
Minzheng Wang, Yongbin Li, Haobo Wang, Xinghua Zhang, Nan Xu, Bingli Wu, Fei Huang, Haiyang Yu, and Wenji Mao. 2025.
\newblock Think on your feet: Adaptive thinking via reinforcement learning for social agents.
\newblock \emph{arXiv e-prints}, pages arXiv--2505.

\bibitem[{Wang et~al.(2024)Wang, Ma, Zhang, Ni, Chandra, Guo, Ren, Arulraj, He, Jiang et~al.}]{wang2024mmlu}
Yubo Wang, Xueguang Ma, Ge~Zhang, Yuansheng Ni, Abhranil Chandra, Shiguang Guo, Weiming Ren, Aaran Arulraj, Xuan He, Ziyan Jiang, and 1 others. 2024.
\newblock Mmlu-pro: A more robust and challenging multi-task language understanding benchmark.
\newblock \emph{Advances in Neural Information Processing Systems}, 37:95266--95290.

\bibitem[{Wang(2025)}]{wang2025geometry}
Zhixiang Wang. 2025.
\newblock The geometry of persona: Disentangling personality from reasoning in large language models.
\newblock \emph{arXiv preprint arXiv:2512.07092}.

\bibitem[{Wei et~al.(2022)Wei, Bosma, Zhao, Guu, Yu, Lester, Du, Dai, and Le}]{wei2022finetuned}
Jason Wei, Maarten Bosma, Vincent Zhao, Kelvin Guu, Adams~Wei Yu, Brian Lester, Nan Du, Andrew~M. Dai, and Quoc~V Le. 2022.
\newblock Finetuned language models are zero-shot learners.
\newblock In \emph{International Conference on Learning Representations}.

\bibitem[{Wei et~al.(2023)Wei, Huang, Lu, Zhou, and Le}]{wei2023simple}
Jerry Wei, Da~Huang, Yifeng Lu, Denny Zhou, and Quoc~V Le. 2023.
\newblock Simple synthetic data reduces sycophancy in large language models.
\newblock \emph{arXiv preprint arXiv:2308.03958}.

\bibitem[{Wu et~al.(2024)Wu, Shi, Rahmani, Ramineni, and Yilmaz}]{wu2024understanding}
Bin Wu, Zhengyan Shi, Hossein~A Rahmani, Varsha Ramineni, and Emine Yilmaz. 2024.
\newblock Understanding the role of user profile in the personalization of large language models.
\newblock \emph{arXiv preprint arXiv:2406.17803}.

\bibitem[{Xiao et~al.(2025)Xiao, Ma, Song, Xu, Zhang, Wang, and Fu}]{xiao2025human}
Changrong Xiao, Wenxing Ma, Qingping Song, Sean~Xin Xu, Kunpeng Zhang, Yufang Wang, and Qi~Fu. 2025.
\newblock Human-ai collaborative essay scoring: A dual-process framework with llms.
\newblock In \emph{Proceedings of the 15th International Learning Analytics and Knowledge Conference}, pages 293--305.

\bibitem[{Yang et~al.(2025)Yang, Li, Yang, Zhang, Hui, Zheng, Yu, Gao, Huang, Lv et~al.}]{yang2025qwen3}
An~Yang, Anfeng Li, Baosong Yang, Beichen Zhang, Binyuan Hui, Bo~Zheng, Bowen Yu, Chang Gao, Chengen Huang, Chenxu Lv, and 1 others. 2025.
\newblock Qwen3 technical report.
\newblock \emph{arXiv preprint arXiv:2505.09388}.

\bibitem[{Zhang et~al.(2025{\natexlab{a}})Zhang, Lin, Hou, Feng, and Li}]{zhang2025adaptthink}
Jiajie Zhang, Nianyi Lin, Lei Hou, Ling Feng, and Juanzi Li. 2025{\natexlab{a}}.
\newblock Adaptthink: Reasoning models can learn when to think.
\newblock \emph{arXiv preprint arXiv:2505.13417}.

\bibitem[{Zhang et~al.(2024)Zhang, Zeng, Hua, Ding, Chen, Ma, Li, Cui, Qi, Zhu, Lv, Hu, Liu, and Zhou}]{ultramedical}
Kaiyan Zhang, Sihang Zeng, Ermo Hua, Ning Ding, Zhang-Ren Chen, Zhiyuan Ma, Haoxin Li, Ganqu Cui, Biqing Qi, Xuekai Zhu, Xingtai Lv, Jin-Fang Hu, Zhiyuan Liu, and Bowen Zhou. 2024.
\newblock \href {https://doi.org/10.52202/079017-0819} {Ultramedical: Building specialized generalists in biomedicine}.
\newblock In \emph{Advances in Neural Information Processing Systems}, volume~37, pages 26045--26081. Curran Associates, Inc.

\bibitem[{Zhang et~al.(2025{\natexlab{b}})Zhang, Wang, Zhang, Li, Song, Li, Qiu, Cao, Cai, Yao et~al.}]{zhang2025leveraging}
Shao Zhang, Xihuai Wang, Wenhao Zhang, Chaoran Li, Junru Song, Tingyu Li, Lin Qiu, Xuezhi Cao, Xunliang Cai, Wen Yao, and 1 others. 2025{\natexlab{b}}.
\newblock Leveraging dual process theory in language agent framework for real-time simultaneous human-ai collaboration.
\newblock \emph{arXiv preprint arXiv:2502.11882}.

\bibitem[{Zhao et~al.(2025)Zhao, Zhang, Wang, Liang, Li, and Wong}]{zhao2025frem}
Zhengyi Zhao, Shubo Zhang, Zezhong Wang, Bin Liang, Binyang Li, and Kam-Fai Wong. 2025.
\newblock Frem: A flexible reasoning mechanism for balancing quick and slow thinking in long-context question answering.
\newblock \emph{arXiv preprint arXiv:2503.22985}.

\end{thebibliography}

\appendix

\section*{Appendix}

\section{Effect of Personalized Information on Model Performance}
\label{app:introduction}

To systematically investigate the impact of personalized information on the performance of large language models, we evaluate different types of LLMs: general-purpose LLMs (DeepSeek-R1 and Qwen3-30B-A3B-Thinking) and personalized LLMs (ALIGNXPERT-ICA and ALIGNXPERT-PBA). The evaluation is conducted on both objective tasks (PubMedQA and TriviaQA) and subjective personalized tasks (PersonaFeedback and FSPO-roleplay). In addition, we design different personalized information input scenarios, including no user information, user background irrelevant to the query, and user background relevant to the query, to examine how different types of personalized information affect model performance. The experimental results are shown in Table \ref{tab:intro}.

\begin{table*}[!ht]
\centering
\renewcommand{\arraystretch}{1.15}
\resizebox{\textwidth}{!}{%
\begin{tabular}{llcccc}
\hline
\multirow{2}{*}{\textbf{Type}} & \multirow{2}{*}{\textbf{Benchmark}} & \multirow{2}{*}{\textbf{Model}} & \multicolumn{3}{c}{\textbf{Persona Setting}}         \\
                               &                                     &                                 & No persona     & Unaligned persona & Aligned persona \\ \hline
\multirow{8}{*}{Objective}     & \multirow{4}{*}{PubMedQA}           & DeepSeek-R1                     & \textbf{0.452} & 0.345             & 0.426           \\
                               &                                     & Qwen3-30B-A3B-Thinking          & 0.444          & 0.385             & \textbf{0.472}  \\
                               &                                     & ALIGNXPERT-ICA            & 0.368          & 0.385             & \textbf{0.480}  \\
                               &                                     & ALIGNXPERT-PBA            & \textbf{0.521} & 0.389             & 0.495           \\
                               & \multirow{4}{*}{TriviaQA}           & DeepSeek-R1                     & \textbf{0.654} & 0.624             & 0.615           \\
                               &                                     & Qwen3-30B-A3B-Thinking          & \textbf{0.585} & 0.563             & 0.580           \\
                               &                                     & ALIGNXPERT-ICA            & \textbf{0.470} & 0.398             & 0.464           \\
                               &                                     & ALIGNXPERT-PBA            & \textbf{0.470} & 0.400             & 0.461           \\ \hline
\multirow{8}{*}{Personalized}  & \multirow{4}{*}{PersonaFeedback}    & DeepSeek-R1                     & 0.520          & 0.502             & \textbf{0.785}  \\
                               &                                     & Qwen3-30B-A3B-Thinking          & 0.515          & 0.506             & \textbf{0.730}   \\
                               &                                     & ALIGNXPERT-ICA            & 0.538          & 0.522             & \textbf{0.609}           \\
                               &                                     & ALIGNXPERT-PBA            & 0.524          & 0.533             & \textbf{0.649}           \\
                               & \multirow{4}{*}{FSPO-roleplay}      & DeepSeek-R1                     & 0.592          & 0.625             & \textbf{0.738}           \\
                               &                                     & Qwen3-30B-A3B-Thinking          & 0.653          & 0.602             & \textbf{0.759}  \\
                               &                                     & ALIGNXPERT-ICA            & 0.590          & 0.535             & \textbf{0.637}  \\
                               &                                     & ALIGNXPERT-PBA            & 0.600          & 0.528             & \textbf{0.662}  \\ \hline
\end{tabular}%
}
\vspace{-8pt}
\caption{The double-edged effect of personalized information}
\label{tab:intro}
\end{table*}

\section{Detailed Implementations}
\label{app:Detailed Implementations}
\subsection{Details on Training Datasets Construction}

\paragraph{SFT Datasets Construction.} For the personalized task, we use all trajectories in Personalized mode. For the objective task, we do not simply force all trajectories into General mode. Instead, we conduct a gain-based selection: a trajectory remains in Personalized mode only if incorporating persona information helps the model generate the correct answer. Otherwise, for samples where persona information is misleading or provides no clear benefit, we generate trajectories in General mode. This strategy aims to allow the model to leverage persona information when it can improve objective question answering, while avoiding factual errors or logical biases that may arise from inappropriate use of persona.

\paragraph{RL Datasets Construction.}

For the RL stage, we continue to sample data from UltraMedical, FLAN, and ALIGNX to construct the RL training set. For the objective datasets UltraMedical and FLAN, we construct the training data with an equal proportion of unaligned personas and aligned personas, to balance the two personalized information settings encountered by the model during training.

\begin{table}[!ht]
\centering
\small
\renewcommand{\arraystretch}{1.2}
\resizebox{\columnwidth}{!}{%
\begin{tabular}{cllcc}
\toprule
\multirow{2}{*}{\textbf{Stage}} 
& \multirow{2}{*}{\textbf{Type}} 
& \multirow{2}{*}{\textbf{Dataset}} 
& \multicolumn{2}{c}{\textbf{Reasoning Trajectory}} \\
\cmidrule(lr){4-5}
& & & \textbf{Objective} & \textbf{Personalized} \\ 
\midrule
\multirow{3}{*}{\textbf{SFT}}
& \multirow{2}{*}{Objective}
& UltraMedical
& 270 & 1{,}730 \\
& 
& FLAN
& 1{,}181 & 2{,}819 \\
& Personalized
& AlignX
& -- & 2{,}000 \\ 
\midrule
\multirow{3}{*}{\textbf{RL}}
& \multirow{2}{*}{Objective}
& UltraMedical
& \multicolumn{2}{c}{3{,}000} \\
& 
& FLAN
& \multicolumn{2}{c}{5{,}998} \\
& Personalized
& AlignX
& \multicolumn{2}{c}{1{,}000} \\ 
\midrule
\end{tabular}}
\vspace{-8pt}
\caption{The dataset composition of the PersonaDual training set PersonaDualMode. To avoid data leakage, samples overlapping with PubMedQA are filtered out from UltraMedical, and samples overlapping with TriviaQA are filtered out from FLAN.}
\label{tab:training_data}
\end{table}

\subsection{Training Details}

The training hyperparameters of PersonaDual are summarized in Table \ref{tab:parameters}. For the RL stage, we set the "\#Rollouts per Sample" to 8, which corresponds to $n=4$ in Equation \ref{eq:prefixsample}. That is, for each sample, we enforce prefix sampling to generate four responses under the objective reasoning mode and four responses under the personalized reasoning mode. In addition, we set the "\#Prefix advantage weight coefficient" to $2.0$, meaning that during gradient updates, the contribution of the prefix tokens is amplified by a factor of two. 
Intuitively, the advantage scores computed for the prefix portion are multiplied by $2.0$.

\begin{table}[!ht]
\centering
\small
\setlength{\tabcolsep}{6pt}
\renewcommand{\arraystretch}{1.15}
\resizebox{\linewidth}{!}{
\begin{tabular}{lc}
\toprule
\textbf{Stage 1 (SFT)} & \textbf{Value} \\
\midrule
Model Initialization                        & Qwen3-8B-Instruct \\
Global Batch Size                           & 16 \\
Peak Learning Rate                          & $5\mathrm{e}{-5}$ \\
Learning Rate Scheduler                     & Cosine \\
Training Epochs                             & 1 \\
Warm-up Ratio                               & 0.01 \\
Max Sequence Length (SFT)                   & 4096 \\
Numerical Precision                         & bfloat16 \\
GPU Usage                                   & 8 NVIDIA A800 \\
DeepSpeed Configuration                     & ZeRO-3 \\
\midrule
\textbf{Stage 2 (RL)} & \textbf{Value} \\
\midrule
Model Initialization                        & Stage 1 \\
Global Batch Size                           & 64 \\
Peak Learning Rate                          & $1\mathrm{e}{-6}$ \\
Learning Rate Scheduler                     & Linear \\
Training Epochs                             & 5 \\
Warm-up Ratio                               & 0.1 \\
Max Prompt Length                           & 2048 \\
Max Response Length                         & 1024 \\
KL Penalty Coefficient                      & 0.04 \\
Generation Temperature                      & 0.6 \\
\# Rollouts per Sample                      & 8 \\
\# Prefix advantage weight coefficient      & 2.0 \\
Numerical Precision                         & bfloat16 \\
GPU Usage                                   & 8 NVIDIA A800 \\
DeepSpeed Configuration                     & ZeRO-3 \\
\bottomrule
\end{tabular}}
\vspace{-8pt}
\caption{Two-stage training hyperparameters of PersonaDual.}
\label{tab:parameters}
\end{table}

\subsection{Evaluation Datasets}
\label{app:benchmarks}

In this work, we evaluate our method on seven datasets (shown in Table \ref{tab:eval_benchmarks}) across two types of scenarios: objective tasks and subjective personalized tasks. Below, we briefly describe the datasets used.

\textbf{MMLU-Pro.} A professional subset of the Massive Multitask Language Understanding benchmark, covering multiple domains such as history, law, and social sciences. It evaluates general reasoning skills using multiple-choice questions.

\textbf{SuperGPQA.} A large-scale question-answering dataset focusing on general knowledge and reasoning across diverse topics.  The questions are presented in multiple-choice format to assess the model’s objective problem-solving ability.

% \begin{table}[H]
% \centering
% \small
% \renewcommand{\arraystretch}{1.4}
% \caption{Evaluation benchmarks for objective and personalized capabilities.}
% \label{tab:eval_benchmarks}
% \vspace{-8pt}
% \resizebox{\columnwidth}{!}{   % 自动缩放宽度到单栏
% \begin{tabular}{cllcc} \hline
% \textbf{Dimension}
% & \textbf{Domain}
% & \textbf{Benchmark}
% & \textbf{Sample Size}
% & \textbf{Output Type} \\ \hline
% \multirow{5}{*}{Objective}
% & General reasoning & MMLU-Pro     & 1{,}000 & Multiple-choice \\
% & General reasoning & SuperGPQA    & 1{,}000 & Multiple-choice \\
% & General reasoning & TriviaQA     & 1{,}000 & Free-form QA \\
% & Medicine          & PubMedQA     & 1{,}000 & Free-form QA \\
% & Math              & MATH500      &   500   & Math problem solving \\ \hline
% \multirow{2}{*}{Personalized}
% & Personalized QA   & PersonaFeedback & 1{,}000 & Multiple-choice \\
% & Personalized QA   & FSPO-roleplay   & 1{,}500 & Multiple-choice \\ \hline
% \end{tabular}
% }
% \end{table}

\begin{table}[t]
\centering
\small
\setlength{\tabcolsep}{5pt}
\renewcommand{\arraystretch}{1.15}
\resizebox{\columnwidth}{!}{
\begin{tabular}{llcc}
\toprule
\textbf{Dim.} & \textbf{Benchmark (Domain)} & \textbf{N} & \textbf{Type} \\
\midrule
\multirow{5}{*}{Obj.}
& MMLU-Pro (General reasoning)     & 1{,}000 & MCQ \\
& SuperGPQA (General reasoning)    & 1{,}000 & MCQ \\
& TriviaQA (General reasoning)     & 1{,}000 & Free-form \\
& PubMedQA (Medicine)              & 1{,}000 & Free-form \\
& MATH500 (Math)                   &   500   & Math sol. \\
\midrule
\multirow{2}{*}{Pers.}
& PersonaFeedback (Personalized QA) & 1{,}000 & MCQ \\
& FSPO-roleplay (Personalized QA)   & 1{,}500 & MCQ \\
\bottomrule
\end{tabular}}
\vspace{-8pt}
\caption{Evaluation benchmarks for objective and personalized capabilities.}
\label{tab:eval_benchmarks}
\end{table}

\begin{figure}[!t]
    \centering
    \includegraphics[width=\columnwidth]{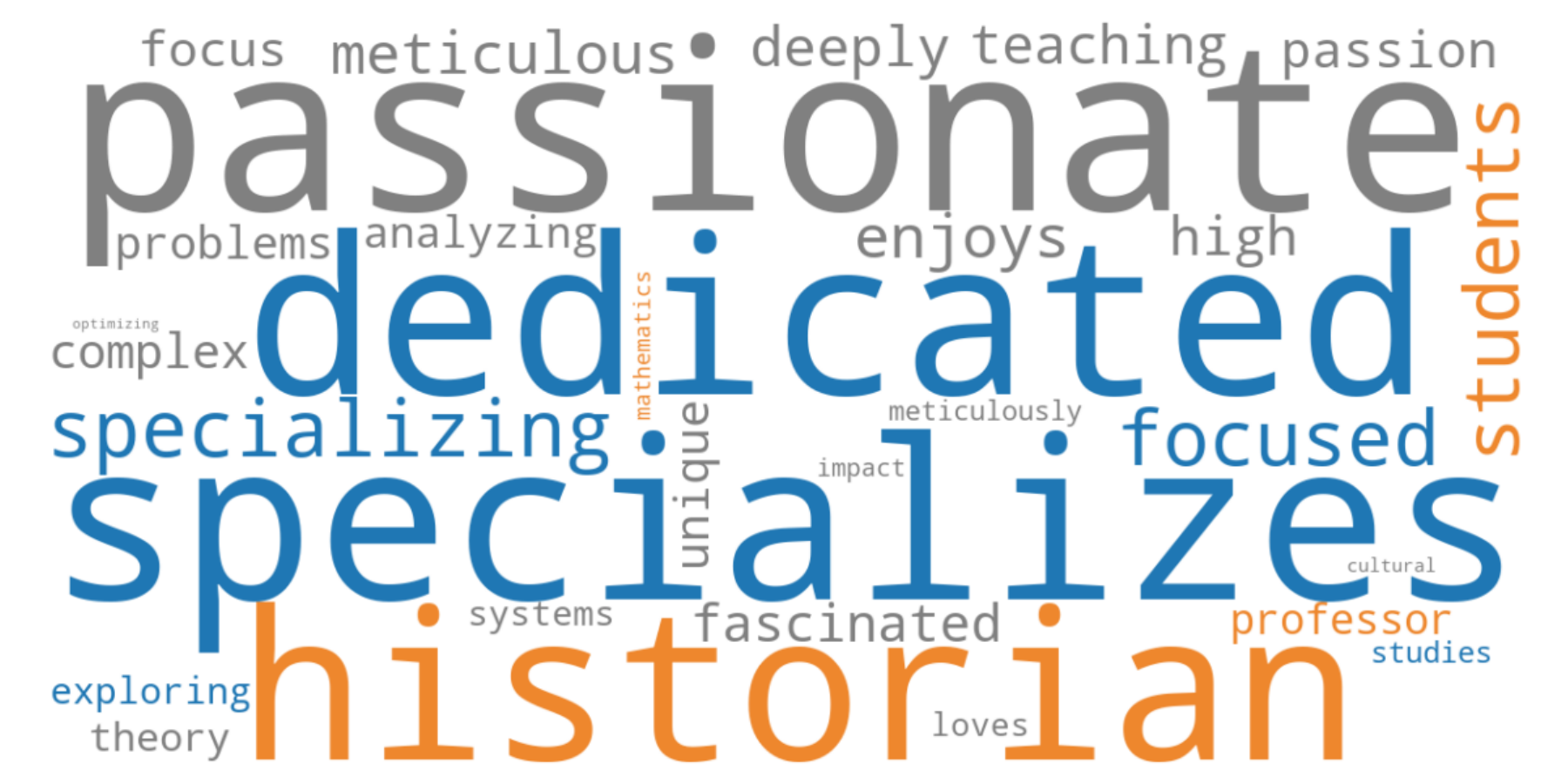}  % 图片路径
    \caption{Extracted keywords from personas}
    \label{fig:keywords}
\end{figure}

\textbf{TriviaQA.} A dataset of open-domain trivia questions, requiring factual knowledge and reasoning.  The questions are free-form, allowing models to generate natural language answers.

\textbf{PubMedQA.} A biomedical question-answering dataset derived from PubMed abstracts, focusing on professional medical knowledge. The questions are free-form, testing the model’s ability to provide concise, domain-specific answers. Since our baselines include Llama3-8B-Instruct, which primarily operates in English, we translate the original Chinese PersonaFeedback dataset into English to ensure a fair and consistent comparison across models.

\textbf{MATH500.} A dataset of mathematical problems designed to evaluate quantitative reasoning and problem-solving skills.  The problems require step-by-step solutions and are presented in free-form.

\textbf{PersonaFeedback.} A personalized QA dataset providing user personas and preference pairs. The model must choose answers that best align with the given persona, using multiple-choice questions.

\textbf{FSPO-roleplay.} Similar to PersonaFeedback, this roleplay-based dataset provides user personas and response preferences. The model selects answers that reflect the persona’s preferences in multiple-choice format.

\section{Further Discussion for PersonaDual Upperbound}
\label{app:upperbound}

Table \ref{tab:upperbound} reports the upper bound performance of PersonaDual, where an instance is considered correct if either the objective reasoning mode or the personalized reasoning mode produces a correct answer. We observe that PersonaDual already surpasses the performance of using a single reasoning mode on most datasets in the table. This indicates that the current adaptive mode selection mechanism is effective. Moreover, the upper bound is substantially higher than that of any single mode, suggesting significant potential for the PersonaDual paradigm.

\begin{table*}[!ht]
\centering
\small
\setlength{\tabcolsep}{4.5pt}
\renewcommand{\arraystretch}{1.2}
\resizebox{\linewidth}{!}{
\begin{tabular}{l|cccccc|ccc}
\toprule
\multirow{2}{*}{\textbf{Model}} &
\multicolumn{6}{c|}{\textbf{Objective Acc. (Unalign./Align.)}} &
\multicolumn{3}{c}{\textbf{Personalized Acc. (Align.)}} \\
\cmidrule(lr){2-7}\cmidrule(lr){8-10}
& \textbf{PubMedQA} & \textbf{TriviaQA} & \textbf{MMLU-Pro} & \textbf{SuperGPQA} & \textbf{MATH500} & \textbf{Avg.}
& \textbf{P.FB.} & \textbf{F.RP.} & \textbf{Avg.} \\
\midrule
PersonaDual-general    & 0.509/0.524 & 0.445/\textbf{0.520} & 0.624/\textbf{0.666} & \textbf{0.332}/0.326 & 0.788/0.802 & 0.540/0.568 & 0.722 & 0.794 & 0.758 \\
PersonaDual-personal   & \textbf{0.517}/0.537 & 0.441/0.518 & \textbf{0.627}/0.649 & 0.322/0.344 & 0.790/0.792 & 0.539/0.568 & 0.740 & \textbf{0.823} & \textbf{0.782} \\
PersonaDual            & 0.508/\textbf{0.549} & \textbf{0.449}/0.515 & 0.622/0.657 & 0.331/\textbf{0.347} & \textbf{0.790}/\textbf{0.808} & \textbf{0.540}/\textbf{0.575} & \textbf{0.747} & 0.799 & 0.773 \\
PersonaDual Upperbound & 0.622/0.613 & 0.493/0.570 & 0.696/0.717 & 0.408/0.415 & 0.855/0.846 & 0.615/0.632 & 0.776 & 0.847 & 0.812 \\
\bottomrule
\end{tabular}}
\vspace{-8pt}
\caption{Upperbound of PersonaDual. Objective benchmarks report accuracy under Unaligned/Aligned persona settings (Unalign./Align.). Personalized benchmarks are evaluated under the Aligned setting. Avg. denotes the mean accuracy across datasets within each block.}
\label{tab:upperbound}
\end{table*}

\section{Additional Analysis of PersonaDual Generalization}
\label{app:llama}
\begin{table*}[!ht]
\centering
\small
\renewcommand{\arraystretch}{1.2}
\resizebox{\textwidth}{!}{
\begin{tabular}{l|cccccc|ccc}
\toprule
\multirow{2}{*}{\textbf{Model}} &
\multicolumn{6}{c|}{\textbf{Objective Acc. (Unalign./Align.)}} &
\multicolumn{3}{c}{\textbf{Personalized Acc. (Align.)}} \\
\cmidrule(lr){2-7}\cmidrule(lr){8-10}
& \textbf{PubMedQA} & \textbf{TriviaQA} & \textbf{MMLU-Pro} & \textbf{SuperGPQA} & \textbf{MATH500} & \textbf{Avg.}
& \textbf{P.FB.} & \textbf{F.RP.} & \textbf{Avg.} \\
% FSPO-roleplay - > FSPO
\midrule

\multicolumn{10}{c}{\cellcolor[RGB]{247, 238, 199}{\textit{General purpose Models}}} \\
LLaMA3.1-8B-Instruct
& 0.400/0.492 & 0.400/0.462 & 0.409/0.460 & 0.187/0.194 & 0.443/0.458 & 0.368/0.413
& 0.586 & 0.613 & 0.600 \\

CoT$^{\ast}$
& 0.507/0.520 & 0.505/0.557 & 0.417/0.461 & 0.215/0.232 & 0.385/0.382 & 0.406/0.430
& 0.737 & 0.750 & 0.744 \\

G-SFT-RL$^{\ast}$
& 0.504/0.536 & \textbf{0.517}/0.553 & 0.423/0.442 & 0.231/0.227 & 0.361/0.374 & 0.407/0.426
& 0.647 & 0.682 & 0.665 \\

\midrule
\multicolumn{10}{c}{\cellcolor[RGB]{240,207,207}{\textit{Personalization-oriented Models}}} \\

Personal-Prompt$^{\ast}$
& 0.422/0.470 & 0.335/0.483 & 0.387/0.462 & 0.152/0.190 & 0.455/0.462 & 0.350/0.413
& 0.590 & 0.617 & 0.604 \\

P-SFT-RL$^{\ast}$
& 0.245/0.439 & 0.436/0.560 & 0.302/0.409 & 0.162/0.205 & 0.193/0.282 & 0.268/0.379
& \textbf{0.738} & 0.712 & 0.725 \\

ALIGNXPERT-ICA$^{\ast}$
& 0.398/0.486 & 0.418/0.465 & 0.320/0.385 & 0.124/0.166 & 0.448/0.456 & 0.342/0.392
& 0.539 & 0.623 & 0.581 \\

ALIGNXPERT-PBA$^{\ast}$
& 0.379/0.477 & 0.425/0.469 & 0.311/0.349 & 0.119/0.162 & \textbf{0.460/0.474} & 0.339/0.386
& 0.531 & 0.621 & 0.576 \\

\midrule
\multicolumn{10}{c}{\cellcolor[RGB]{198,230,204}{\textit{Dual-mode Models}}} \\

PersonaDual-Prompt$^{\ast}$
& 0.362/0.414 & 0.494/0.494 & 0.357/0.375 & 0.146/0.192 & 0.453/0.462 & 0.362/0.387
& 0.556 & 0.625 & 0.591 \\

PersonaDual-Router$^{\ast}$
& 0.496/0.437 & \textbf{0.517/0.561} & 0.414/0.412 & 0.228/0.208 & 0.355/0.292 & 0.402/0.382
& 0.736 & 0.721 & 0.729 \\

PersonaDual$^{\ast}$
& \textbf{0.527/0.538} & \textbf{0.517}/0.553 & \textbf{0.434/0.468} & \textbf{0.245/0.251} & 0.405/0.413 & \textbf{0.426/0.445}
& \textbf{0.738} & \textbf{0.751} & \textbf{0.745} \\

\midrule
No-Persona Upperbound
& 0.542 & 0.530 & 0.445 & 0.245 & 0.410 & 0.434
& -- & -- & -- \\

\bottomrule
\end{tabular}}
% \vspace{-8pt}
\caption{\textbf{Bold} indicates the best performance on each benchmark (evaluated under the corresponding setting). Models marked with ${\ast}$ are trained (or constructed via prompting) based on LLama3.1-8B-Instruct.}
\label{tab:llamaresult}
\end{table*}

To examine the generality of the proposed PersonaDual framework, we replace the backbone with LLaMA3.1-8B-Instruct and repeat the main experiment, with results reported in Table \ref{tab:llamaresult}. Overall, PersonaDual consistently achieves the best performance (objective: 0.426/0.445, personalized: 0.745) among all compared methods, demonstrating that the framework generalizes well beyond other baselines.

On objective benchmarks, PersonaDual shows strong robustness under persona-unaligned settings, achieving an average accuracy that is only 0.008 lower than the no-persona upper bound, indicating that the framework largely avoids detrimental persona interference. Under persona-aligned settings, PersonaDual further improves objective performance by 0.011 on average, highlighting its ability to effectively leverage persona information when alignment is beneficial.

We observe that PersonaDual, as well as several trained baselines, exhibits a performance drop on MATH500 compared to the backbone model. This degradation may be related to differences in how the training data aligns with the pre-trained knowledge structure of Llama models. Despite this challenge, PersonaDual remains the best-performing method among all models trained under the same conditions.

Taken together, these results indicate that the PersonaDual framework is broadly applicable across different backbone models, providing both strong objective performance and effective personalization without sacrificing robustness under persona mismatch.

\end{document}